\documentclass{article}

\usepackage[main, final]{neurips_2026}

\usepackage[utf8]{inputenc} 
\usepackage[T1]{fontenc}    
\usepackage{hyperref}       
\usepackage{url}            
\usepackage{booktabs}       
\usepackage{amsfonts}       
\usepackage{nicefrac}       
\usepackage{microtype}      
\usepackage{xcolor}         
\usepackage{graphicx}
\usepackage{amsmath}
\usepackage{wrapfig}
\usepackage{multirow}
\usepackage{siunitx}

\title{
    \includegraphics[height=1.5em]{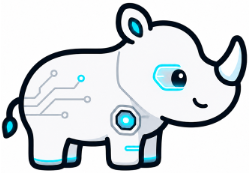}
    \quad RhinoVLA Technical Report
}

\author{%
  Huixi Technology
}

\begin{document}

\maketitle

\begin{abstract}

Vision-Language-Action (VLA) models have shown strong potential for robotic
manipulation, but real-time deployment on edge hardware remains
challenging. In this work, we identify VLM visual
and context tokens as a major source of deployment latency: for GEMM-dominated
projection operators, computation grows linearly with the number of input tokens
when model dimensions are fixed. Motivated by this observation, we propose
RhinoVLA, a deployment-oriented VLA model co-designed with the Huixi R1 edge
SoC. RhinoVLA adopts a token-efficient Qwen3-VL backbone and a continuous
Action Expert, reducing the VLM-side token and computation burden while
preserving pretrained multimodal capability. To support cross-robot learning,
RhinoVLA further introduces a unified interface that combines View Registry,
72D physical state-action slot space, and robot-instance LoRA,
allowing heterogeneous robot observations and action schemas to be aligned
under a shared policy. On the
deployment side, RhinoVLA is optimized through hardware-aware
compilation, mixed-precision execution, and parallel visual encoding. Experiments
show that RhinoVLA achieves downstream performance comparable to $\pi_{0.5}$ at a similar parameter scale, while reaching 11.69 Hz end-to-end inference on Huixi R1,
meeting the 10 Hz real-time closed-loop control target. The project will be
open-sourced at \url{https://github.com/HuixiAI/RhinoVLA}.

\end{abstract}

\section{Introduction}

Vision-Language-Action (VLA) models have recently shown strong potential for
robotic manipulation and embodied decision-making~\cite{brohan2022rt1,zitkovich2023rt2,kim2025openvla,black2024pi0,bjorck2025grootn1}. By combining visual
perception, language understanding, and continuous action generation, these
models provide a promising path toward general-purpose robot policies. However,
deploying VLA models on real robots remains difficult. Modern VLA systems often
use pretrained vision-language backbones with long
multimodal contexts, and iterative action-generation modules. These components
improve policy capability, but they also introduce substantial computation and
memory traffic, making real-time closed-loop inference challenging on onboard
edge hardware.

The key deployment bottleneck is not model size alone, but how the VLA token
structure maps to edge hardware. In a typical VLA pipeline, the VLM backbone
processes visual observations and language context before the Action Expert
generates continuous actions. Our analysis of $\pi_{0.5}$~\cite{pi05technicalreport} shows that the VLM
Backbone and Action Expert dominate end-to-end latency on NVIDIA Jetson AGX
Orin~\cite{nvidia2022jetsonorin}, together accounting for more than 90\% of the total runtime. A closer
operator-level breakdown further shows that most VLM latency comes from MLP
projection operators. Since these projections are GEMM-dominated, their FLOPs
scale linearly with the number of VLM visual and context tokens when the hidden
dimensions are fixed. Therefore, two VLA models with a similar parameter scale
can have very different inference speeds if they use different visual-token
organizations.

This token-scaling property directly motivates our algorithm-system
co-design. On the algorithm side, we need a VLM backbone that is not only
pretrained for strong multimodal reasoning, but also efficient in its visual
token representation. We therefore build RhinoVLA on Qwen3-VL~\cite{bai2025qwen3vl}. Under a common
$256\times256$ image setting, Qwen3-VL can represent one image with about 64
merged visual tokens, while PaliGemma-224~\cite{beyer2024paligemma} uses 256 image tokens. This lower
visual-token cost is especially important for VLA inference, where multiple
camera views and language context are processed by the VLM before action
generation. On the hardware side, because the VLM occupies a large fraction of
VLA runtime and its dominant operators are mostly compute-bound, VLA
inference benefits from an edge SoC with high onboard compute headroom. We
therefore deploy RhinoVLA on Huixi self-developed chip R1, a 7 nm edge SoC with 500 TOPS INT8
compute, providing stronger AI compute headroom than Orin-class edge platforms.
To translate this peak capability into actual inference speed, we further
optimize RhinoVLA on R1 with hardware-aware compilation, mixed-precision
execution, and parallel visual encoding.

Beyond inference efficiency, RhinoVLA is designed to address another central
challenge in VLA training: heterogeneous robot datasets do not share a unified
input-output interface. Different robots may use different camera layouts,
view orders, sensing modalities, action definitions, end effectors, and
low-level control conventions. Directly pooling such data can make the learning
problem ill-defined, because the same image index may correspond to different
camera roles, and the same action dimension may denote different physical
quantities across robots. RhinoVLA resolves this mismatch with a unified
interface and a unified pre-training strategy. A View Registry explicitly tags
each image with its camera role and modality. A unified 72D physical slot
space, together with binary state and action masks, assigns fixed physical
semantics to state-action dimensions and excludes invalid slots from
supervision. Robot-instance LoRA~\cite{hu2022lora} modules model residual embodiment-specific
behavior without introducing robot-specific output heads. During pre-training,
RhinoVLA jointly optimizes the VLM LoRA, shared Action Expert, and
robot-instance LoRA on mixed robot data. The masked flow-matching loss
supervises only valid action slots, while residual regularization keeps the
LoRA adapters focused on embodiment-specific corrections. This lets RhinoVLA
learn shared visuomotor structure while retaining low-cost robot adaptation.

\begin{figure}[ht]
    \centering
    \includegraphics[width=0.95\linewidth]{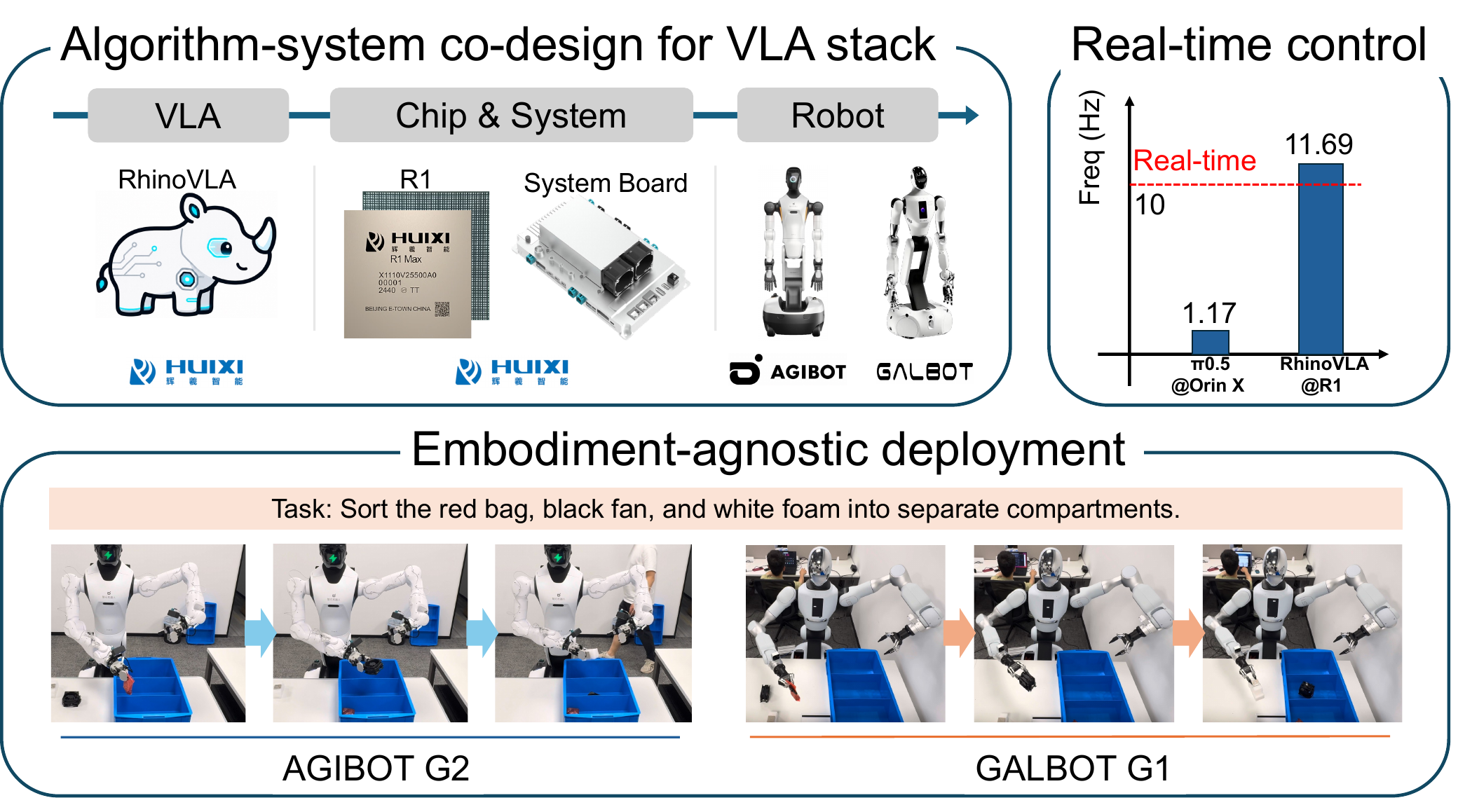}
    \caption{RhinoVLA achieves real-time edge control at 11.69 Hz through algorithm–system co-design and enables embodiment-agnostic deployment across diverse robots.}
    \label{fig:overview}
\end{figure}

Shown in Figure~\ref{fig:overview}, these algorithm--deployment co-design choices translate into a VLA
system with both competitive policy performance and real-time onboard
inference. RhinoVLA keeps a parameter scale comparable to $\pi_{0.5}$, but
achieves substantially higher end-to-end speed by reducing the VLM token burden
and improving hardware utilization on R1. It reaches downstream task accuracy
comparable to $\pi_{0.5}$ and runs at 11.69 Hz on Huixi R1, meeting the 10 Hz
closed-loop inference target~\cite{luo2024hilserl,jang2022bcz}.

Our contributions are summarized as follows:

\begin{itemize}
    \item We identify VLM visual and context tokens as a key source of VLA
    deployment cost. Based on this observation, we design RhinoVLA around a
    token-efficient Qwen3-VL backbone, enabling faster inference than
    PI-style models at a comparable parameter scale. 

    \item We introduce a unified cross-robot training framework. This framework enables RhinoVLA to learn shared visuomotor structure across
    heterogeneous robot embodiments while preserving embodiment-specific
    flexibility.

    \item Through algorithm-system co-optimization, RhinoVLA achieves downstream task accuracy comparable to $\pi_{0.5}$ while reaching 11.69 Hz end-to-end inference on the Huixi R1 edge SoC.

\end{itemize}
\section{Background}

\subsection{Vision-Language-Action Models}

Early VLA models mainly followed an end-to-end robot policy design. RT-1~\cite{brohan2022rt1} uses a transformer-based policy that takes visual observations and language instructions as inputs and predicts tokenized robot actions, trained mainly on large-scale robot trajectories from the Everyday Robots platform. RT-2~\cite{zitkovich2023rt2} further extends this paradigm by representing robot actions as text tokens and co-fine-tuning a pretrained VLM on both robot trajectories and web-scale vision-language data, making robot control compatible with next-token prediction. However, in these systems, visual-language understanding and action prediction are still tightly coupled within a single policy, and the action interface is largely discretized or tokenized for scalable training.

Recent VLA models have gradually shifted toward a more modular architecture. Systems such as $\pi_{0}$~\cite{black2024pi0}, $\pi_{0.5}$~\cite{pi05technicalreport}, and GR00T N1~\cite{bjorck2025grootn1} typically build on pretrained VLM backbones for visual-language understanding, while introducing dedicated action modules for continuous robot control. $\pi_{0}$ uses a flow-matching action expert on top of a pretrained VLM; $\pi_{0.5}$ combines discrete action-token pretraining with continuous flow-matching post-training; and GR00T N1 adopts a dual-system design where a VLM module interprets visual-language inputs and a DiT-based~\cite{peebles2023scalable} action module generates motor actions. Meanwhile, training data has expanded from relatively narrow robot-platform datasets toward cross-embodiment and heterogeneous mixtures, including multi-robot trajectories, web vision-language data, simulation, synthetic data, and human videos. This evolution improves the task-level capability and cross-embodiment generalization of VLA models, but also makes edge deployment increasingly challenging due to the high computation cost of pretrained VLM backbones, long multimodal context, and the real-time latency requirements of closed-loop robot control. Therefore, designing more efficient VLA architectures is necessary for practical deployment on resource-constrained robotic platforms.

\subsection{VLM Backbones for VLA Models}

The development of VLA backbones follows the broader progress of
vision-language models. Early VLMs mainly focused on large-scale image-text
alignment: CLIP~\cite{radford2021clip} learned transferable visual representations from natural-language
supervision, while later models such as Flamingo and LLaVA~\cite{liu2023llava} extended VLMs toward
interleaved image-text understanding, few-shot visual prompting, and visual
instruction following. As a result, pretrained VLMs have become a natural choice
for providing semantic visual-language representations in robot policies.

When used in VLA models, the VLM backbone must balance semantic capability and
deployment cost. Large backbones can improve instruction understanding and
generalization, but their parameter size and visual token cost directly affect
robotic deployment, especially on edge hardware. Existing VLA systems therefore
span different scales. RT-2~\cite{zitkovich2023rt2} uses large PaLI-family VLMs~\cite{beyer2024paligemma}, with RT-2-X~\cite{openx2023rtx} reaching
the 55B scale, and represents robot actions as text tokens. OpenVLA~\cite{kim2025openvla} uses a 7B
Prismatic-style VLM built from DINOv2~\cite{oquab2023dinov2}, SigLIP~\cite{zhai2023sigmoid}, and Llama~2~\cite{touvron2023llama}, and is trained on
970k robot demonstrations. In contrast, $\pi_{0}$ adopts a more compact
PaliGemma backbone of about 3B parameters and adds a 300M-parameter action
expert, resulting in a 3.3B-parameter VLA. This comparison shows a practical
trend: modern VLA systems increasingly use a pretrained VLM as the semantic
backbone, but smaller and more efficient backbones are preferred when real-time
edge deployment is considered.

The Qwen-VL series~\cite{wang2024qwen2vl,qwen2025qwen25vl,bai2025qwen3vl} is a promising backbone direction under this trend. Earlier
Qwen-VL variants have already been used in VLA systems, such as DexVLA~\cite{wen2025dexvla} with
Qwen2-VL-2B and LingBot-VLA~\cite{wu2026pragmatic} with Qwen2.5-VL. Qwen3-VL further improves several
capabilities that are directly useful for VLA: spatial-temporal modeling,
fine-grained visual feature usage, long interleaved multimodal context, and
timestamp-based video grounding. These upgrades better match multi-view robot
observations, short visual histories, object grounding, and temporally changing
scenes. Recent Qwen3-VL-based VLA examples, including VLA Foundry and
InternVLA-A1, suggest that Qwen3-VL is becoming an important candidate backbone
for future efficient VLA systems.

\subsection{Edge Hardware for VLA Deployment}
Existing VLA deployment hardware mainly includes desktop GPUs and NVIDIA Jetson edge platforms. Desktop GPUs such as RTX 4090/5090~\cite{nvidia_geforce_rtx4090_specs,nvidia_geforce_rtx5090_specs} offer strong compute capability and a mature CUDA ecosystem, but their high power, large form factor, and cooling requirements make them unsuitable for on-board robotic deployment. Jetson platforms such as Orin~\cite{nvidia2022jetsonorin} and Thor~\cite{nvidia2025jetsonthor_datasheet} are more compact and widely used in robotics; however, Orin has limited compute headroom for 10 Hz VLA inference, while Thor offers higher performance at a much higher system cost.

Recently, high-performance Chinese edge SoCs from vendors such as Huixi have also been adopted by robotics companies. In this work, we use Huixi R1, a 7 nm SoC designed for embodied intelligence. R1 provides 500 TOPS INT8 compute, an 8-core SIMT architecture, and 200 GB/s-class memory bandwidth, offering strong compute and memory support for multimodal visual encoding and VLA inference. It has also been adopted by leading embodied intelligence companies such as AgiBot as an on-board compute chip.
\section{Method}
\subsection{Performance analysis}
\subsubsection{VLA Roofline Analysis}
To quantify the deployment bottleneck of current VLA models on edge hardware, we conduct an end-to-end roofline analysis using NVIDIA Jetson AGX Orin as a representative platform. Orin provides a theoretical FP16 throughput of approximately 43 TFLOPS and a memory bandwidth of around 203 GB/s. Considering practical factors such as operator efficiency, scheduling overhead, irregular memory access, and model complexity, we assume an ideal compute utilization of 40\%, corresponding to an effective FP16 throughput of approximately 17.2 TFLOPS.

\begin{figure}[ht]
    \centering
    \includegraphics[width=0.75\linewidth]{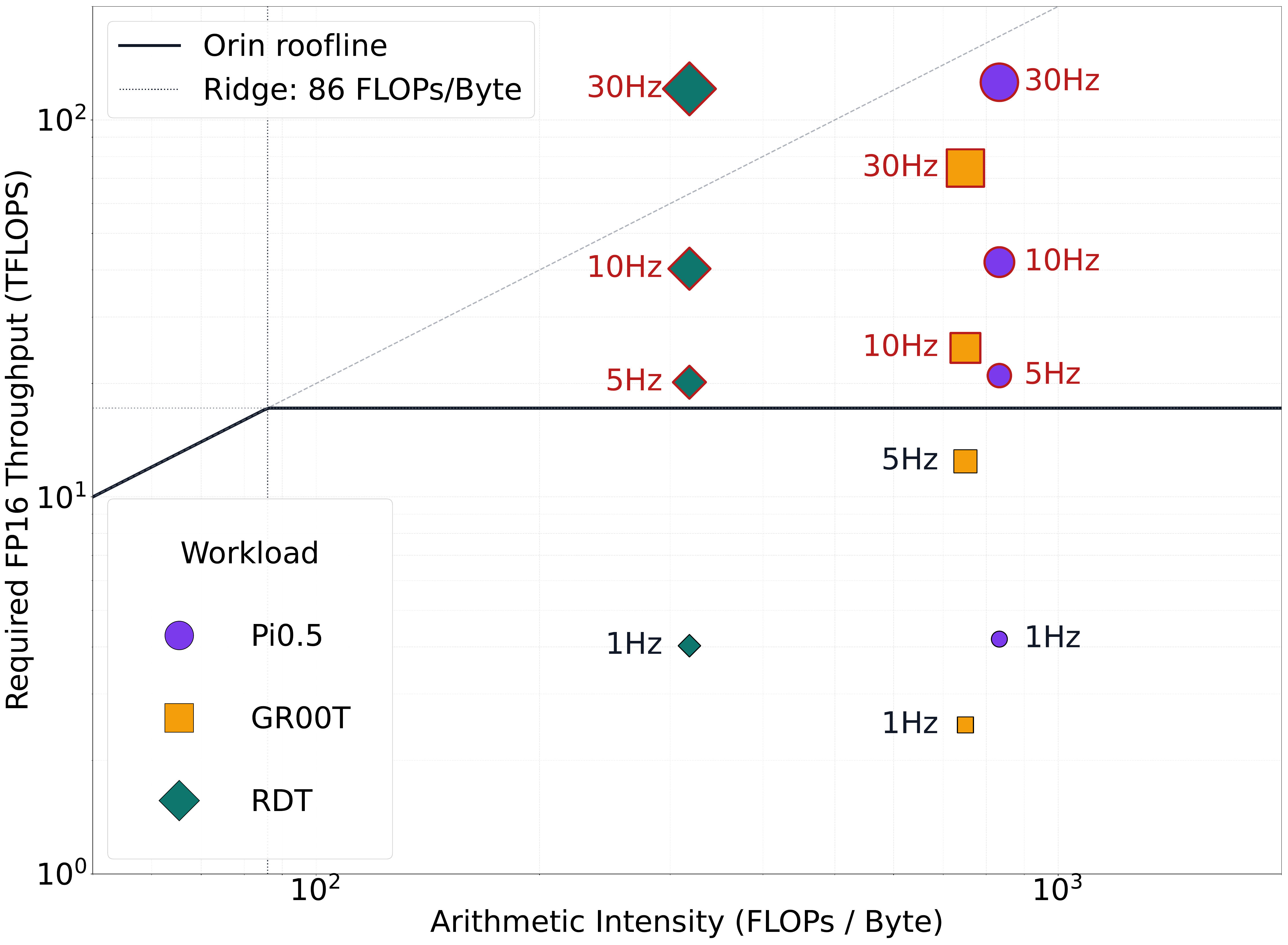}
    \caption{End-to-end roofline analysis of representative VLA models on NVIDIA Jetson AGX Orin under FP16 precision.}
    \label{fig:vla_roofline_orin}
\end{figure}

We select several representative VLA models, including $\pi_{0.5}$, GR00T N1, and RDT~\cite{pi05technicalreport,bjorck2025grootn1,liu2024rdt}, and estimate their end-to-end computational requirements under different control frequencies. The estimation covers the main inference stages, including vision encoding, VLM prefill, and action expert inference. As shown in  Figure~\ref{fig:vla_roofline_orin}, $\pi_{0.5}$ and RDT already approach or exceed Orin’s effective roofline limit at a target frequency of 5 Hz, and significantly exceed the hardware capability at 10 Hz and above. This indicates that, even under the relatively optimistic assumption of 40\% compute utilization, current edge hardware can hardly support these VLA models to meet the minimum real-time closed-loop control requirement of 10 Hz.

\subsubsection{VLA Latency breakdown}
To identify the major sources of latency in end-to-end VLA inference, we conduct a breakdown analysis of the $\pi_{0.5}$ PyTorch-SDPA inference pipeline on NVIDIA Jetson AGX Orin. The results show that, within the total end-to-end latency of approximately 858.3 ms, the vision encoder, VLM Backbone, and action expert take 69.3 ms, 528.0 ms, and 257.0 ms, respectively. Among them, the VLM Backbone and action expert together account for more than 90\% of the end-to-end inference time, making them the dominant performance bottlenecks in the $\pi_{0.5}$ inference pipeline. Therefore, to meet the real-time closed-loop control requirement at the 10 Hz level, the latency introduced by the VLM Backbone and action expert should be the primary optimization targets.

To further clarify the optimization direction for the VLM Backbone, we take $\pi_{0.5}$ as an example and perform an operator-level latency breakdown of its VLM component on the Orin platform. 
The results show that the latency of the VLM is not evenly distributed across all operators, but is highly concentrated in the MLP module inside the Transformer blocks. Specifically, the three linear projection operators, gate\_proj, up\_proj, and down\_proj, together account for approximately 74.7\% of the VLM latency. In contrast, the attention projection operators, including Q\_proj, K\_proj, V\_proj, and O\_proj, account for only about 7.2\% in total, while the remaining operators account for approximately 18.1\%. This indicates that, during $\pi_{0.5}$ VLM inference, the MLP module is a more dominant source of latency than the attention projections.

From the roofline perspective, gate\_proj, up\_proj, and down\_proj are essentially GEMM operators with relatively high arithmetic intensity. For a typical linear layer, given an input matrix

\begin{equation}
X \in \mathbb{R}^{B \times S \times d_{\text{in}}}
\label{eq:input_shape}
\end{equation}

and a weight matrix

\begin{equation}
W \in \mathbb{R}^{d_{\text{in}} \times d_{\text{out}}}
\label{eq:weight_shape}
\end{equation}

the output can be written as

\begin{equation}
Y = XW,\quad Y \in \mathbb{R}^{B \times S \times d_{\text{out}}}
\label{eq:linear_projection}
\end{equation}

The computational cost can be approximated as

\begin{equation}
\text{FLOPs} = 2 B S d_{\text{in}} d_{\text{out}}.
\label{eq:gemm_flops}
\end{equation}

where B denotes the batch size, S denotes the number of input tokens, and $d_{in}$ and $d_{out}$ denote the input and output channel dimensions, respectively. For gate\_proj, up\_proj, and down\_proj in the MLP module, such computation is repeatedly performed in every Transformer block. Therefore, the total computational cost grows linearly with the number of input tokens, hidden size, intermediate size, and the number of layers.

As shown by the formula, when the weight dimensions are largely fixed, the computational cost of MLP projection operators is proportional to the number of input tokens S. Therefore, the key to reducing VLM Backbone latency is to reduce the input size to the VLM, especially the number of visual tokens and context tokens. Based on this observation, one design principle of RhinoVLA is to compress visual tokens, remove redundant context, and optimize the organization of multimodal tokens, thereby reducing the MLP GEMM computation at the source, alleviating the compute pressure on edge devices, and lowering the overall VLM latency.

\subsection{RhinoVLA Architecture}
\paragraph{Overview.}
RhinoVLA follows the two-module VLA decomposition used by $\pi_{0.5}$: a
visual-language backbone encodes robot observations and language instructions,
and an Action Expert generates continuous action chunks with flow matching.
This design preserves a strong pretrained VLM for perception and instruction
understanding, while keeping action generation in a separate robot-control
module.

\smallskip
\noindent
For the visual-language backbone, the latency analysis motivates reducing the
token burden at the VLA model level. Since the dominant MLP projection
operators scale linearly with the number of visual and context tokens,
RhinoVLA adopts a 2.13B-parameter Qwen3-VL~\cite{bai2025qwen3vl} as its
visual-language backbone. Under the common $224\times224$ image setting,
Qwen3-VL represents one image with 64 merged visual tokens after spatial
merging, whereas the PaliGemma-224~\cite{beyer2024paligemma} backbone used by
$\pi_{0.5}$ uses 256 image tokens. This reduces the visual token burden by
4$\times$ before the multimodal sequence enters the language backbone. For
multi-view VLA inputs, where several camera streams are processed together
with language instructions, this reduction directly lowers the MLP GEMM
computation analyzed above. At the same time, Qwen3-VL provides strong
pretrained multimodal capability, making it a suitable backbone for
token-efficient VLA inference without weakening visual-language reasoning.

\smallskip
\noindent
The Action Expert keeps a comparable scale to $\pi_{0.5}$, with 0.40B
parameters versus about 0.43B, but is redesigned around the Qwen3-VL interface
rather than directly reusing the original Gemma expert. It follows
Qwen-compatible transformer components, including attention/cache handling and
text-MLP blocks. At each denoising step, the Action Expert conditions on the
last 18 layers of the Qwen3-VL KV cache, the current 72D robot state,
state/action masks, the noisy action chunk, the flow-matching time, and the
robot-instance index. It predicts a flow velocity over the unified 72D action
slot space, with invalid dimensions masked out. In this way, action generation
uses reusable visual-language context from the VLM, while the state/action
masks and robot-instance index expose the robot-specific control interface to
the Action Expert.

\begin{figure}[t]
    \centering
    \includegraphics[width=0.98\linewidth]{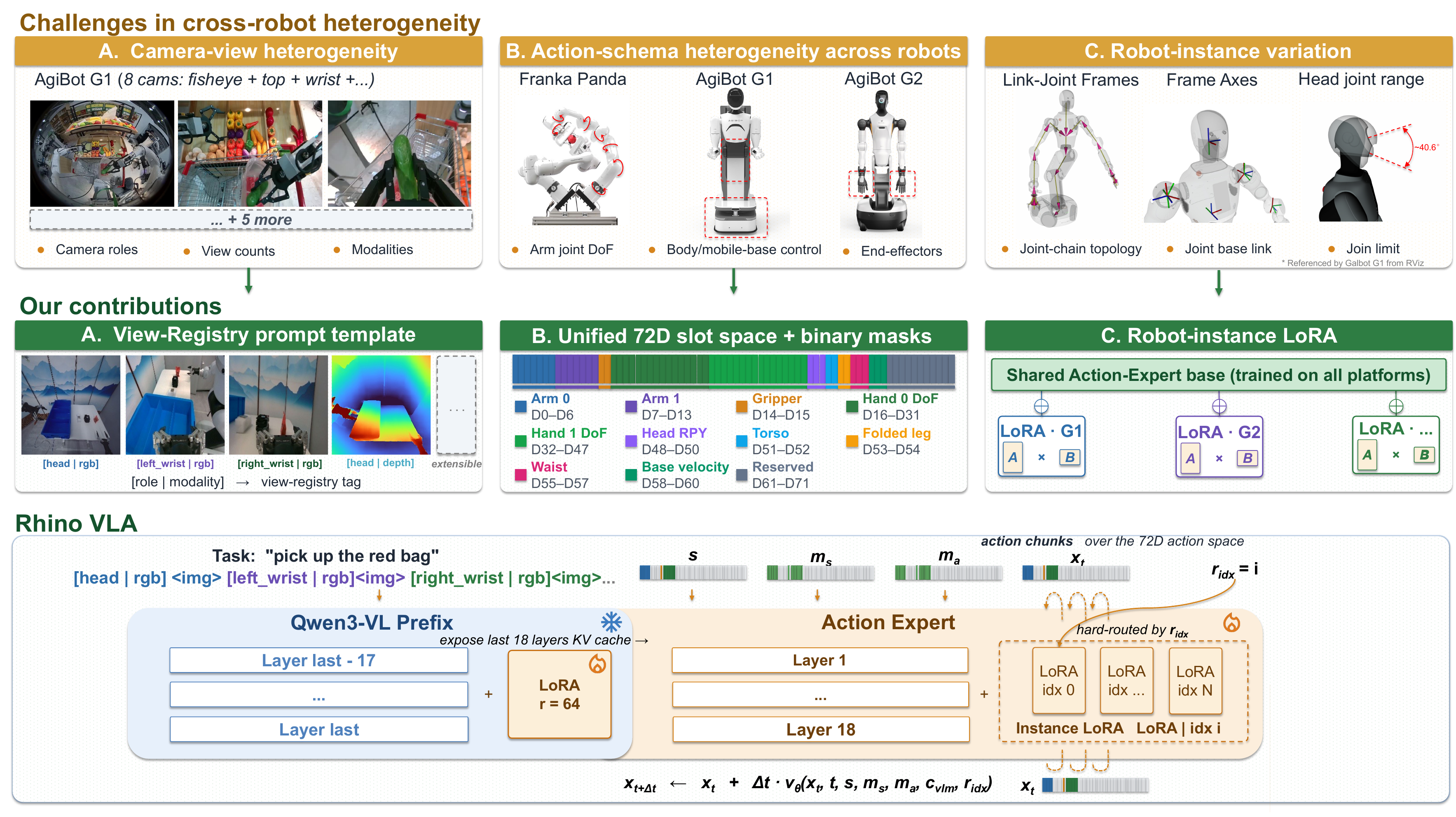}
    \caption{Overview of RhinoVLA. The architecture aligns heterogeneous
    robot datasets through three interface mechanisms. The Action Expert conditions on Qwen3-VL
    visual-language memory and predicts masked flow velocities over active
    action slots.}
    \label{fig:Rhino_vla_architecture}
\end{figure}

\smallskip
\noindent
While the compact VLM improves efficiency, cross-robot training introduces
heterogeneity in observations, action interfaces, and robot embodiments. To
enable a shared policy across diverse platforms, RhinoVLA incorporates three
key mechanisms (Figure~\ref{fig:Rhino_vla_architecture}): a view registry for
observation alignment, a unified 72D physical state--action slot space for
action alignment, and robot-instance LoRA adapters for embodiment-specific
adaptation. We detail these components next.

\paragraph{Challenge A: Camera-view heterogeneity.}
Robot datasets often use different camera layouts and naming conventions.
Although many datasets contain comparable view types, such as front,
head-mounted, overhead, or wrist cameras, they may differ in camera count,
mounting pose, field name, view order, and modality. Without explicit view
labels, the model must infer camera identity from image order or
dataset-specific field names. This is unreliable in cross-robot training: the
first image in one dataset may be a head view, whereas the first image in
another dataset may be a wrist view.

\paragraph{Mechanism A: View Registry.}
RhinoVLA introduces a View Registry that maps each dataset-specific camera
field to a fixed role-modality vocabulary during preprocessing. The
corresponding tag is inserted before the image content, such as
\texttt{[head|rgb]}, \texttt{[left\_wrist|rgb]}, or \texttt{[head|depth]}.
As a result, Qwen3-VL receives both the image content and its explicit camera
identity before tokenization, while different samples may still contain
different numbers of views.

The View Registry decouples camera identity from dataset-specific image order.
It makes camera observations comparable across datasets while preserving
view-dependent cues: head or front cameras usually provide stable global
context, whereas wrist cameras provide close-up manipulation views that move
with the arm. Explicit view tags prevent these distinctions from being hidden
inside dataset-specific conventions. Table~\ref{tab:Rhino_vla_view_registry}
summarizes the View Registry vocabulary and prompt template.

\begin{table}[t]
\centering
\small
\begin{tabular}{p{0.26\linewidth} p{0.64\linewidth}}
\hline
\textbf{Prompt field} & \textbf{Example / value} \\
\hline
Task text &
\texttt{pick up the cup} \\
View role &
\texttt{front}, \texttt{left\_wrist}, \texttt{head} \\
Modality &
\texttt{rgb}, \texttt{depth}, \texttt{rgbd} \\
Image content &
structured Qwen-VL image item \\
\hline
\multicolumn{2}{p{0.90\linewidth}}{
\textbf{Training prompt template:}
\newline
\texttt{Task: \{task\}}
\newline
\texttt{Views:}
\newline
\texttt{[\{role\_1\} | \{modality\_1\}] + image\_1}
\newline
\texttt{[\{role\_2\} | \{modality\_2\}] + image\_2}
\newline
\texttt{...}
\newline
\texttt{[\{role\_n\} | \{modality\_n\}] + image\_n}
} \\
\hline
\end{tabular}
\caption{View Registry prompt fields and training template. Each image is
explicitly tagged with its camera role and modality, allowing heterogeneous
camera layouts to be represented through a shared prompt interface.}
\label{tab:Rhino_vla_view_registry}
\end{table}

\paragraph{Challenge B: Action-schema heterogeneity.}
Robot action vectors cannot be shared by length or index position alone. The
same vector index may refer to different physical quantities in different
datasets, and some robots may not expose the corresponding degree of freedom at
all. Directly pooling such action vectors would assign physically different
meanings to the same output dimension, making cross-robot action learning
ill-defined.

\paragraph{Mechanism B: Unified 72D slot space with binary masks.}
RhinoVLA uses a unified 72D slot space as the shared state-action contract
across robots. Each slot has a fixed physical meaning across datasets. State
and action use the same slot definition: the state records the current physical
value, while the action label represents the future target for the same slot
whenever such a target is defined. For mobile-base slots, the action is defined
as a velocity target rather than a position target. A dataset may activate only
a subset of slots, but it cannot redefine their semantics.

Binary masks are attached to the same 72D space. The state mask indicates
which state dimensions exist in the current sample, and the action mask
indicates which output dimensions are valid for supervision. As model inputs,
the masks tell the Action Expert which slots are present in the current robot.
During training, invalid or physically absent action slots are excluded from
the flow-matching supervision. This prevents missing dimensions from being
treated as zero-valued targets and avoids introducing spurious supervision for
robot-specific degrees of freedom.

The 72D contract follows the motivation of physically interpretable action
spaces inspired by RDT~\cite{liu2024rdt}. Active output dimensions are tied to physical quantities:
arm, wrist, head, and waist angles use radians; parallel grippers use a closed
ratio in $[0,1]$; and base commands use metric velocity units. These slot
groups define the physical coordinate system exposed to the Action Expert, as
listed in Table~\ref{tab:Rhino_vla_72d_slots}. Detailed mapping rules,
including dexterous-hand allocation and converter constraints, are provided in
Appendix~\ref{app:slot_mapping_details}.

\begin{table}[t]
\centering
\scriptsize
\begin{tabular}{p{0.14\linewidth} p{0.26\linewidth} p{0.18\linewidth} p{0.32\linewidth}}
\hline
\textbf{Slots} & \textbf{Physical group} & \textbf{Unit} & \textbf{Examples} \\
\hline
D0--D6 &
Arm 0 canonical joints &
rad &
upper arm, forearm, and wrist joints for a single arm or left arm \\
D7--D13 &
Arm 1 canonical joints &
rad &
right arm in a bimanual robot \\
D14--D15 &
Parallel grippers &
closed ratio &
0 is open, 1 is closed \\
D16--D31 &
Hand 0 active DoF &
rad &
active hand-control slots for the first hand \\
D32--D47 &
Hand 1 active DoF &
rad &
active hand-control slots for the second hand \\
D48--D50 &
Head / neck RPY &
rad &
roll, pitch, and yaw when available \\
D51--D52 &
Torso pitch / lift &
rad, m &
torso pitch and vertical lift \\
D53--D54 &
Folded-leg mechanism joints &
rad &
two actuated joints for foldable leg mechanisms \\
D55--D57 &
Waist RPY &
rad &
waist roll, pitch, and yaw \\
D58--D60 &
Base velocity command &
m/s, m/s, rad/s &
base $v_x$, $v_y$, and yaw rate \\
D61--D71 &
Reserved auxiliary slots &
reserved &
inactive in the current schema \\
\hline
\end{tabular}
\caption{Unified 72D slot space for RhinoVLA. The table defines the shared
physical coordinate system used to align heterogeneous robot state and action
schemas. Binary masks specify which slots are valid for each robot and sample.}
\label{tab:Rhino_vla_72d_slots}
\end{table}

Together, the slot definitions and binary masks form a single robot interface:
slots fix the physical meaning of each dimension, while masks select the subset
that is valid for a particular robot and sample. This allows RhinoVLA to
train on heterogeneous robots without forcing all datasets to share the same
action-vector length or index convention.

\paragraph{Challenge C: Robot-instance residual variation.}
The View Registry and unified 72D slot space align camera observations and
state-action semantics, but they do not remove all embodiment-specific
differences. Two robots may share the same nominal action slots while still
responding differently due to calibration errors, joint limits, gripper
mechanics, camera placement, payload, low-level controllers, or action scaling.
A single shared policy must therefore capture both cross-robot common structure
and instance-specific residual behavior.

\paragraph{Mechanism C: Robot-instance LoRA.}
RhinoVLA places robot-instance LoRA modules inside the Action Expert. The
shared Action Expert receives gradients from all datasets and learns common
visuomotor structure across robot platforms. For each sample, the corresponding
LoRA module is selected by \texttt{instance\_id} and learns a small
robot-specific correction on top of the shared base model.

We use robot-instance LoRA instead of robot-specific output heads for two
reasons. First, separate output heads would weaken the unified 72D action
contract by allowing each robot to learn its own final action mapping. Second,
they would introduce robot-specific deployment graphs, making it harder to
reuse the same inference kernels and hardware-specific operator optimizations
across robots. In contrast, robot-instance LoRA adapts the internal
action-generation features while keeping the attention modules, final action
projection, and 72D output interface shared.

The LoRA modules are inserted into the feed-forward network of every Action
Expert layer, while the attention modules and final action projection remain
shared. This placement keeps the main action-generation computation common
across robots, while providing enough capacity to model residual
instance-specific behavior. During deployment, the selected robot-instance LoRA
can be merged into the base Action Expert weights.

This design provides three practical advantages:
\begin{enumerate}
    \item \textbf{Unified deployment graph.} After merging the selected
    robot-instance LoRA, every robot uses the same 18-layer Action Expert
    structure. The deployed model does not require robot-specific computation
    graphs, so the same inference kernels and hardware-specific operator
    optimizations can be reused.

    \item \textbf{Sparse adapter activation.} Although the training code may
    reserve multiple instance ids, each sample is hard-routed by
    \texttt{instance\_id} to one robot-instance LoRA. A forward pass therefore
    activates only the selected adapter, not all reserved adapters.

    \item \textbf{Low-cost robot extension.} Adding a new robot does not
    require a new action head or changes to the unified 72D slot space, as long
    as the robot can be mapped to existing slots. The new robot only requires a
    small robot-instance LoRA and its normalization statistics.
\end{enumerate}

\subsection{Training Strategy}

This section describes how the architecture above is trained. We first convert
heterogeneous robot demonstrations into the shared visual and action interface,
then train the policy with unified cross-embodiment pretraining, and finally
adapt the pretrained model to a target real robot with a small amount of task
data.

\subsubsection{Training Data and Standardization}

The training mixture is assembled from multiple robot demonstration sources,
including joint-space subsets from Open X-Embodiment~\cite{openx2023rtx}, and AgiBotWorld~\cite{bu2025agibotworld}. These sources cover single-arm and dual-arm robots,
parallel grippers, dexterous hands, and multiple manipulation tasks.

Each sample is converted into the interface in
Figure~\ref{fig:Rhino_vla_architecture} before it reaches the model. Camera
fields are mapped to View Registry. Native robot states and actions are
mapped to the 72D physical slot space. State and action masks record which
slots are valid for the current robot. The 72D space reserves 16 slots for each
hand. These slots follow a 4-3-3-3-3 allocation from thumb to little finger:
four active DoF for the thumb and three active DoF for each other finger. Hand
actions are mapped to this layout according to the active-joint semantics,
joint order, units, and limits specified by the corresponding hardware or SDK
manual. This allocation matches the design pattern used by many five-finger
dexterous hands, while lower-DoF hands can be represented by masking the
missing finger joints. (Only motor-actuated active joints occupy hand slots;
passive or mechanically coupled joints are not treated as valid action
dimensions.)

Only samples that can be reliably mapped to the 72D physical slot space and
its corresponding masks are used for training. The current pretraining model
does not supervise leg or foot joints, so those fields are excluded from the
training targets when they appear in the source data.

\subsubsection{Pre-training}

RhinoVLA is trained as a cross-embodiment policy on the
mixed robot demonstration corpus. The Qwen3-VL backbone is kept frozen, while
three trainable components are optimized together: the VLM LoRA, the shared
Action Expert, and the robot-instance LoRA inside the Action Expert. The VLM
LoRA adapts the visual-language backbone to robot camera views and manipulation
instructions. The shared Action Expert learns the common 72D action-generation
policy across robots. The selected instance LoRA is routed by
\texttt{instance\_id}.

Training batches are drawn from the mixed robot datasets with the same
power-law balancing rule used in $\pi$-style VLA training,
\[
p_i = \frac{N_i^{0.43}}{\sum_j N_j^{0.43}},
\]
where $N_i$ is the number of training samples in dataset $i$. This gives larger
datasets higher sampling probability without allowing them to dominate every
batch. State and action masks follow the 72D slot convention defined above.
They specify which physical slots are available for the current robot, and only
active action slots contribute to the flow-matching objective.

For a clean target chunk $z \in \mathbb{R}^{H \times 72}$, Gaussian noise
$a \sim \mathcal{N}(0,I)$, and interpolation time $t \in [0,1]$, we construct
\[
x_t = (1-t)a + t z .
\]
The Action Expert predicts
\[
\hat{v}_{\theta} =
f_{\theta}(x_t, t, s, m_s, m_a, c_{\mathrm{vlm}}, r),
\]
where $s$ denotes the 72D robot state, $m_s$ and $m_a$ are the state and action
masks, and $c_{\mathrm{vlm}}$ denotes the last-18-layer Qwen3-VL KV cache used
as visual-language conditioning for the Action Expert. The variable $r$ selects
the robot-instance LoRA used by the current sample. The flow target is $z-a$,
and the main masked flow-matching loss is
\[
\mathcal{L}_{\mathrm{FM}} =
\frac{\sum_{h,d} m_a(d) w(h,d)
\left\|\hat{v}_{\theta}(h,d) - (z(h,d)-a(h,d))\right\|_2^2}
{\sum_{h,d} m_a(d) w(h,d) + \epsilon},
\]
where $w(h,d)$ is a per-slot weighting term used to balance action groups.
During training, the base Action Expert prediction is also kept under direct
supervision and the adapter residual is regularized. This prevents the
instance LoRA from taking over the full action-generation problem: the shared
Action Expert remains responsible for the cross-robot policy, while the
instance LoRA captures smaller deviations caused by robot-specific hardware and
control behavior.

\subsubsection{Post-training and Real-Robot Transfer}

After pretraining, RhinoVLA can be adapted to a target real-robot task without
changing the model interface. The pretrained model already provides shared
visual-language conditioning, the unified 72D state-action schema, and
mask-conditioned action generation. Therefore, transfer only requires fitting
the target-task behavior within the same View Registry, physical slot space,
masks, and robot-instance routing.

During post-training, the visual-language encoder, including the pretrained VLM
LoRA, is frozen. Most Action Expert parameters are also frozen, and we update
only the target robot's instance LoRA. This reuses the same residual adaptation
mechanism for task transfer: the shared base preserves the
cross-robot policy learned during pretraining, while the target instance LoRA
absorbs task- and robot-specific residual corrections from a small real-robot
dataset.

\subsection{Efficient deployment on Huixi R1}
\label{sec:deployment}

\begin{figure}[ht]
    \centering
    \includegraphics[width=0.85\linewidth]{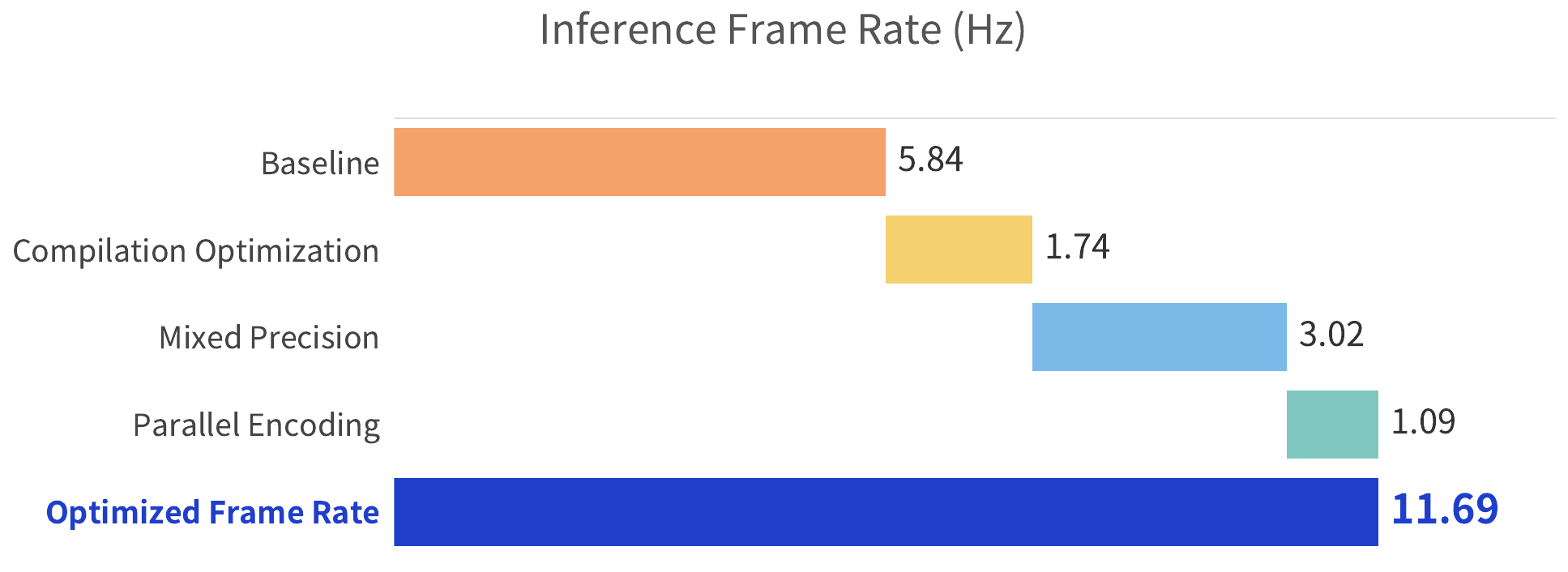}
    \caption{Cumulative frame-rate improvement of RhinoVLA on Huixi R1. Bars for compilation, mixed precision, and parallel encoding denote incremental gains over the previous configuration.}
    \label{fig:framerate}
\end{figure}

We deploy RhinoVLA on our Huixi R1.
Figure~\ref{fig:framerate} reports the runtime frequency of RhinoVLA on the R1 platform after applying a series of deployment optimizations, including compilation optimization, mixed-precision deployment, and parallel encoding. These techniques are implemented with consideration of the hardware characteristics of R1, enabling an end-to-end execution frequency of 11.69 Hz. In the following sections, we describe each optimization technique in detail and analyze its contribution to the overall acceleration.

\textbf{Compilation optimization. }
We optimize the compilation pipeline from three aspects: operator-level optimization, graph-level optimization, and runtime-level optimization.

At the operator level, we optimize the Attention kernels in VLA models according
to the memory hierarchy and execution model of the R1 chip. VLA inference
contains long multimodal sequences composed of language tokens, visual tokens,
robot states, and action-related tokens, making Attention sensitive to off-chip
memory traffic and on-chip data reuse. We therefore adapt FlashAttention-style
tiling~\cite{dao2022flashattention} to R1's software-managed on-chip scratchpad memory (SPM), keeping Q/K/V
tiles, online softmax statistics, and partial outputs on chip whenever possible.
Compared with conventional per-SM shared-memory organization on GPUs, R1 provides
a larger globally coordinated on-chip memory space, which enables larger
attention tiles and more efficient cross-compute-unit reuse and reduction. With
these optimizations, the VLA Attention kernel reaches over 80\% of the theoretical
peak compute throughput on representative input shapes.

At the graph level, we design aggressive operator fusion for VLA Transformer
blocks. In the fused Attention and MLP layers, adjacent operations such as
normalization, linear projection, bias addition, activation, and residual
connection are combined into larger execution units. Intermediate activations
and small parameters such as RMSNorm~\cite{zhang2019root} weights and biases are kept in SPM as much
as possible, while off-chip traffic is mainly dominated by large GEMM weights and
boundary activations. This fused dataflow avoids repeatedly materializing
intermediate tensors in DDR, thereby alleviating the memory-wall bottleneck and
improving energy efficiency by replacing expensive external-memory accesses with
low-cost on-chip SPM accesses.

At the runtime scheduling level, we introduce fine-grained operator task scheduling.
VLA inference contains heterogeneous operators with different compute and memory
characteristics, including GEMM, normalization, activation, data layout
transformation, and softmax. Instead of treating each kernel as a large
homogeneous workload, the R1 runtime decomposes operators into finer-grained
tasks and assigns them according to compute demand, SPM usage, and data-movement
requirements. This allows compute-intensive and memory-sensitive tasks to be
scheduled more flexibly, reducing compute-unit stalls caused by small or
memory-bound operators and improving overall chip utilization.

\textbf{Mixed-precision deployment.} 
In practical deployment, we adopt a mixed-precision quantization scheme with INT8 weights and FP16 activations. Our experiments show that directly applying W8A8 quantization noticeably degrades task success rate and action prediction accuracy. Therefore, we quantize only the model weights to INT8 while keeping activations in FP16. This design reduces weight storage and memory bandwidth cost while avoiding the accuracy loss caused by low-precision activation quantization.

However, W8A16 is not automatically efficient on edge hardware. A conventional implementation usually dequantizes INT8 weights into FP16 before GEMM computation, introducing additional data movement, temporary storage, and conversion overhead. These costs can offset the performance benefits of weight quantization. To address this issue, we implement a customized W8A16 GEMM kernel for the R1 architecture, where weight loading, dequantization, and matrix multiplication are fused into a single execution pipeline.

The kernel design combines the memory bandwidth benefit of quantization with the multi-core parallelism of R1. For major linear layers in the Transformer, GEMM is decomposed into multiple sub-matrix tasks and distributed across eight compute cores, allowing each core to perform relatively independent weight loading and local matrix computation. To reduce DDR channel contention under multi-core execution, we reorganize the offline weight layout in a memory-channel-aware manner, leading to more balanced data streams across physical DDR bandwidth. Within each core, the kernel avoids a two-stage “dequantize-then-compute” execution flow. Instead, weight loading, scale conversion, and matrix multiply-accumulate are organized as an overlapped pipeline, where different processing units handle memory access, INT8-to-FP16 conversion, and FP16 MAC concurrently. We use per-channel scaling to reduce weight storage and bandwidth requirements while preserving the numerical accuracy of linear layers. 

We further conduct kernel-level profiling on key operators in the $\pi_{0.5}$ VLM backbone. For the latency-dominant up\_proj operator, the W16A16 implementation takes 191 $\mu$s, while the customized W8A16 GEMM reduces the latency to 113 $\mu$s, achieving a 1.69× speedup. The kernel also reaches 50.6\% compute utilization. These results show that the proposed W8A16 GEMM implementation effectively reduces memory pressure in linear layers and hides much of the dequantization overhead through fused execution, thereby improving kernel efficiency on R1 while maintaining model accuracy.

\textbf{Parallel encoding. }
RhinoVLA takes three image streams as visual inputs, including one head camera and two wrist cameras mounted on the left and right hands. In the open-source implementation of $\pi_{0.5}$, these images are processed sequentially by the visual encoder. However, we find that this sequential execution is inefficient on R1, because each single-image ViT inference has a relatively low arithmetic intensity and only achieves a low compute utilization. As a result, running the three views one by one leads to poor hardware occupancy and higher latency.

To improve visual encoding efficiency, we replace the sequential execution of three individual ViT forwards with a batched parallel encoding scheme. Specifically, the three camera images are packed into a single batch and processed together by the vision encoder. This increases the effective workload size of each kernel launch, improves data reuse and parallel occupancy on R1. 
Experimental results show that the latency of processing the three input images is reduced from 34.52~ms to 24.31~ms.
\section{Experiment}
\label{sec:exp}

This section evaluates RhinoVLA from training behavior to downstream task
performance. We first verify that unified pre-training is stable, then analyze
whether instance LoRA captures embodiment residuals and improves action
prediction. We then report task-level results in simulation and on real robots.
This order links the final success rates to the training procedure and design
choices that produce them.

\subsection{Setup}
\label{sec:exp:setup}

\paragraph{Evaluation goals.}
The experiments are organized around three questions. First, we examine whether
instance LoRA improves held-out action prediction under the shared 72D action
interface. Second, we ask whether cross-embodiment pretraining improves
downstream policy learning under a standard simulation benchmark. Third, we
test whether the same pretrained policy can be transferred to real robots with
different bodies and end effectors using a small amount of target-task data.

\paragraph{Training data and common model interface.}
RhinoVLA is trained on real-robot trajectories from AgiBotWorld Beta~\cite{contributors2024agibotworldrepo} and AgiBotWorld 2026~\cite{agibotworld2026}, covering 2,976.4 hours on G1 robots and hundreds more on G2 robots, providing both scale and embodiment diversity.

\subsection{Model evaluation}
\label{sec:exp:model}

\begin{figure}[t]
  \centering
  \includegraphics[width=0.98\linewidth]{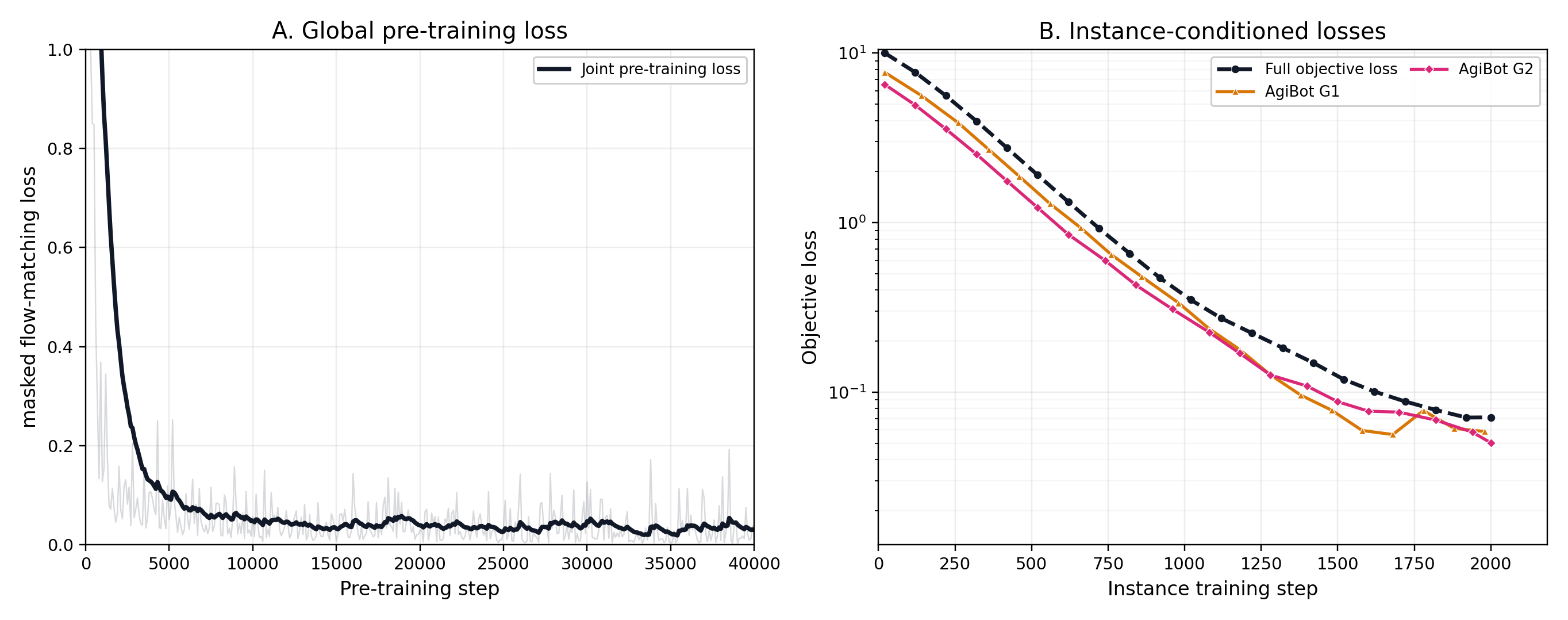}
  \caption{Pre-training loss diagnostics. The left panel shows the global
  masked flow-matching loss, while the right panel reports the full objective
  loss and per-instance losses on AgiBot G1 and AgiBot G2.}
  \label{fig:pretrain_loss}
\end{figure}

\paragraph{Pre-training diagnostics.}
We analyze the masked flow-matching losses from unified pre-training, where the
VLM LoRA, shared Action Expert, and robot-instance LoRA are optimized jointly.
Figure~\ref{fig:pretrain_loss} reports the global/full-objective losses and the
per-instance losses for AgiBot G1 and AgiBot G2. The G1 and G2 curves show
different magnitudes and decay rates, reflecting embodiment- and
controller-specific residuals between the two platforms. These residuals are
handled by the robot-instance LoRA modules, while the shared Action Expert learns
from the mixed cross-robot trajectories through a unified state--action
interface.

\paragraph{Instance LoRA learns embodiment residuals.}
We examine whether the robot-instance LoRA adapters capture
embodiment-specific residuals. For manipulator instances with different action
supports, we compare action-mask Hamming distances with LoRA-residual
similarities measured on a shared probe set.

\begin{figure}[ht]
  \centering
  \includegraphics[width=0.98\linewidth]{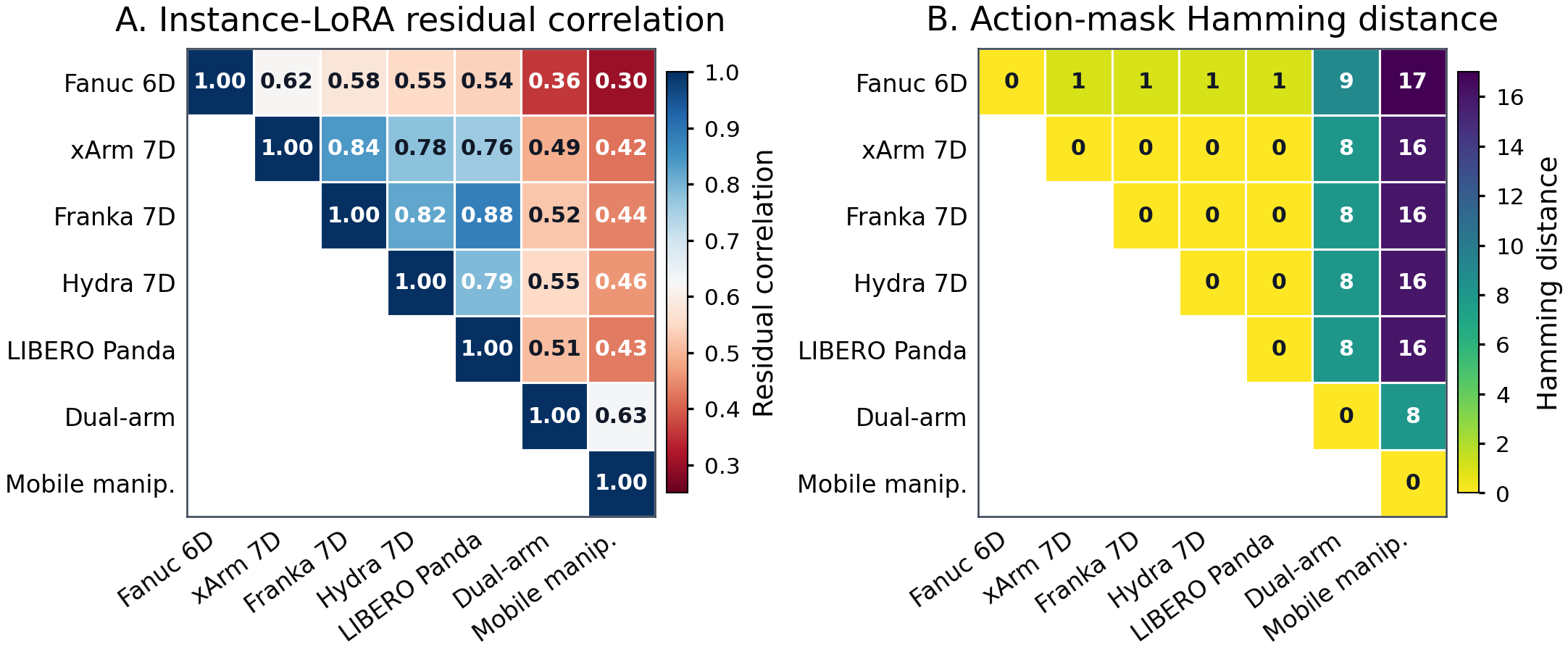}
  \caption{Small-scale diagnostic comparing instance-LoRA residual similarity
  with action-mask distance. The left matrix reports residual correlations
  between instance adapters on the same probe set, while the right matrix
  reports Hamming distances between their active action masks. A smaller mask
  distance corresponds to more similar controllable joint structure.}
  \label{fig:lora_similarity}
\end{figure}

Figure~\ref{fig:lora_similarity} shows that robot pairs with closer action
supports generally have more similar LoRA residuals, whereas structurally
distant pairs show weaker residual similarity. This correspondence indicates
that the adapters encode embodiment-dependent corrections rather than only
dataset identities.

\paragraph{Instance LoRA improves action prediction.}
Table~\ref{tab:ablation} compares the shared Action Expert without instance
LoRA and the full RhinoVLA model under the same validation protocol. Instance
LoRA gives small but consistent gains in masked flow-matching loss,
arm-joint MAE, yaw-rate MAE, and gripper MAE, with little change in
base-velocity MAE. This matches the role of instance adaptation: arm and
gripper dimensions carry most of the embodiment variation in the current robot
mixture, while the shared Action Expert already models the remaining dimensions
well.

\begin{table}[ht]
  \centering
  \small
  \caption{Instance-LoRA ablation on held-out action prediction. Both rows use
  the same validation split, seed, masks, and per-instance dimension weights.
  Masked FM Loss is computed by globally merging the numerator and denominator
  across evaluation shards. MAE is computed only over active slots selected by
  the action mask.}
  \label{tab:ablation}
  \begin{tabular}{lccccc}
    \toprule
    Method & Masked FM Loss & Arm MAE & Base velocity MAE & Yaw rate MAE & Gripper MAE \\
    \midrule
    Base & 0.0192 & 0.0446 & 0.0187 & 0.0195 & 0.1064 \\
    \textbf{Instance LoRA} & \textbf{0.0191} & \textbf{0.0440} & \textbf{0.0188} & \textbf{0.0194} & \textbf{0.1056} \\
    \bottomrule
  \end{tabular}
\end{table}
\paragraph{Simulation Evaluation.}
We evaluate our method on the four standard LIBERO suites: Spatial, Object, Goal, and Long. As shown in Table~\ref{tab:libero_results}, our method achieves an average success rate of 94.1\% with a single jointly trained checkpoint. It outperforms all direct policy baselines and surpasses representative VLA models such as OpenVLA, CoT-VLA, $\pi_0$-FAST, and $\pi_0$. The improvement is particularly evident on the challenging Long suite, where our method reaches 90.4\%, outperforming $\pi_0$ and $\pi_0$-FAST by 17.4 and 30.2 percentage points, respectively.We further observe that robot pretraining improves the average success rate from 90.0\% to 91.8\%, while the proposed View Register design provides an additional gain of 2.3 percentage points, resulting in the final 94.1\% performance. Although this setting still leaves a performance gap to $\pi_{0.5}$, the competitive results against several VLA baselines indicate that a compact pretrained VLM can provide an effective initialization for continuous robot policy learning under limited robot-specific pretraining. Moreover, the gains from the proposed View Register suggest that explicitly modeling camera-view semantics further improves multi-view manipulation performance.


\begin{table}[ht]
\centering
\caption{
Success rates (\%) on the LIBERO benchmark.
S, O, G, and L denote LIBERO-Spatial, LIBERO-Object, LIBERO-Goal, and LIBERO-Long, respectively.}
\label{tab:libero_results}
\small
\setlength{\tabcolsep}{4.8pt}
\renewcommand{\arraystretch}{1.08}
\begin{tabular}{
  @{}l
  S[table-format=2.1]
  S[table-format=2.1]
  S[table-format=2.1]
  S[table-format=2.1]
  S[table-format=2.1]
  @{}
}
\toprule
\textbf{Model} & {\textbf{S}} & {\textbf{O}} & {\textbf{G}} & {\textbf{L}} & {\textbf{Avg}} \\
\midrule

\multicolumn{6}{@{}l}{\textit{Baseline Direct Policy Models}} \\
Diffusion Policy~\citep{chi2025diffusion} & 78.3 & 92.5 & 68.3 & 50.5 & 72.4 \\
Octo~\citep{octo2024}                    & 78.9 & 85.7 & 84.6 & 51.5 & 75.1 \\
MDT~\citep{reuss2024mdt}                  & 78.5 & 87.5 & 73.5 & 64.8 & 76.0 \\

\midrule
\multicolumn{6}{@{}l}{\textit{Baseline VLA Models}} \\
TraceVLA~\citep{zheng2025tracevla}        & 84.6 & 85.2 & 75.1 & 54.1 & 74.8 \\
OpenVLA~\citep{kim2025openvla}            & 84.7 & 88.4 & 79.2 & 53.7 & 76.5 \\
SpatialVLA~\citep{qu2025spatialvla}       & 88.2 & 89.9 & 78.6 & 55.5 & 78.1 \\
WorldVLA~\citep{cen2025worldvla}          & 85.6 & 89.0 & 82.6 & 59.0 & 79.1 \\
CoT-VLA~\citep{zhao2025cotvla}            & 87.5 & 91.6 & 87.6 & 69.0 & 83.9 \\
$\pi_{0}$-FAST~\citep{pertsch2025fast}    & 96.4 & 96.8 & 88.6 & 60.2 & 85.5 \\
$\pi_{0}$~\citep{black2024pi0}            & 90.0 & 86.0 & 95.0 & 73.0 & 86.0 \\
NORA~\citep{hung2025nora}                 & 92.2 & 95.4 & 89.4 & 74.6 & 87.9 \\
SmolVLA~\citep{shukor2025smolvla}         & 93.0 & 94.0 & 91.0 & 77.0 & 88.8 \\
$\pi_{0.5}$~\citep{pi05technicalreport}   & \textbf{98.8} & 98.2 & \textbf{98.0} & \underline{92.4} & \textbf{96.9} \\

\midrule
\multicolumn{6}{@{}l}{\textit{Ours}} \\

RhinoVLA$_{Scratch}$                  & 93.0 & 91.0 & 93.4 & 82.4 & 90.0 \\
\textbf{RhinoVLA}$_{\textbf{Pretrain}}^{-\textbf{VR}}$     & 92.0 & \underline{98.4} & 89.2 & 87.8 & 91.8 \\
\textbf{RhinoVLA}$_{\textbf{Pretrain}}^{+\textbf{VR}}$     & \underline{96.8} & \textbf{99.6} & \underline{97.0} & \textbf{93.0} & \underline{96.6} \\

\bottomrule
\end{tabular}
\end{table}

\paragraph{Real-robot Evaluation.}

Finally, we evaluate whether the same pretrained policy can be adapted to real
robots with different embodiments. The real-robot tasks are conducted on three
commercial robot platforms with distinct controllable structures: \textbf{AgiBot G1}, \textbf{AgiBot G2}, and \textbf{Galbot G1}.

The deployment tasks cover single-arm and bimanual manipulation, short-horizon and long-horizon tasks, as well as rigid and deformable object manipulation. Specifically, Galbot G1 performs multiple short-horizon single-arm pick-and-place tasks, AgiBot G2 performs long-horizon single-arm pick-and-place tasks, and AgiBot G1 performs a bimanual towel-folding task.

We evaluate the models under two deployment settings: seen and unseen. In the seen setting, the adaptation data and evaluation rollouts are collected on the same robot and in the same workspace. In the unseen setting, the evaluation differs from the adaptation data in terms of robot instance, workspace, or object placement distribution. For each task, we conduct multiple trials and report the success rate (SR). The evaluation details and success criteria for different tasks are detailed in Appendix~\ref{sec:Evaluation}. 

\begin{table}[ht]
  \centering
  \small
  \caption{Real-robot success rates after target-task adaptation. SR denotes the percentage of successful real-robot rollouts. The manipulation figures are shown in Figure~\ref{fig:overview} and Figure~\ref{fig:fold}. “-” denotes not evaluated.}
  \label{tab:real_robot_success}
  \begin{tabular}{p{0.14\linewidth} p{0.30\linewidth} p{0.12\linewidth} cc}
    \toprule
    Robot & Task & Setting & $\pi_{0.5}$ SR & RhinoVLA SR  \\
    \midrule
    \multirow{3}{*}{Galbot~G1} 
      & Red bag to the far bin & Unseen & 100\% & 100\%  \\
      & Black fan to the middle bin & Unseen & - & 40\%  \\
      & White foam to the near bin & Unseen & - & 20\%  \\
    \midrule
    \multirow{2}{*}{AgiBot~G2} 
      & \multirow{2}{=}{Three-step sequence} & Seen   & 87\% & 83\%  \\
      &  & Unseen & 52\% & 47\%  \\
    \midrule
    \multirow{2}{*}{AgiBot~G1} 
      & \multirow{2}{=}{Towel folding} & Seen & - & 67\%  \\
      &  & Unseen & - & 43\%  \\
    \bottomrule
  \end{tabular}
\end{table}

Table~\ref{tab:real_robot_success} summarizes the task success rates across different embodiments.
On Galbot G1, an embodiment not covered in pretraining, RhinoVLA shows a certain degree of generalization in the unseen setting. For the task of picking the red bag and placing it into the far bin, RhinoVLA achieves a success rate of 100\%, matching $\pi_{0.5}$.
On AgiBot G2, the evaluation involves more challenging long-horizon pick-and-place tasks, where the robot must understand the instruction and place the red bag, black fan, and white foam into their corresponding bins. RhinoVLA achieves a success rate of 58\% in the seen setting and 24\% in the more difficult unseen setting, outperforming $\pi_{0.5}$ by 6\% and demonstrating stronger generalization.
On AgiBot G1, RhinoVLA achieves success rates of 67\% and 43\% in the seen and unseen settings, respectively, on the towel-folding task. These results further demonstrate RhinoVLA’s effectiveness in deformable object manipulation.

\begin{figure}[ht]
    \centering
    \includegraphics[width=0.95\linewidth]{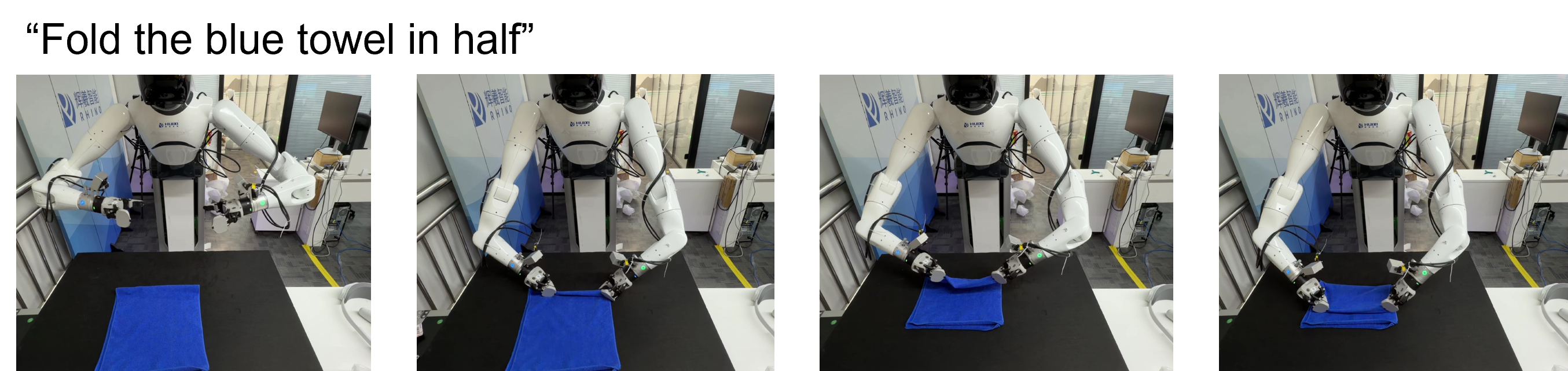}
    \caption{RhinoVLA performs a bimanual towel-folding task on AgiBot G1, demonstrating robustness in deformable object manipulation.}
    \label{fig:fold}
\end{figure}

\subsection{Inference efficiency}
\label{sec:exp:eff}

\paragraph{End-to-end performance on R1.}
After applying the compilation, mixed-precision, and parallel-encoding
optimizations described in Sec.~\ref{sec:deployment}., the full RhinoVLA pipeline runs
end-to-end at \textbf{11.69~Hz} on the Huixi~R1 SoC, up from
5.84~Hz at the unoptimized baseline (Figure~\ref{fig:framerate}). This satisfies
the 10~Hz minimum requirement for real-time closed-loop control.
Table~\ref{tab:e2e} reports the per-stage latency breakdown of the
optimized pipeline.

\begin{table}[ht]
  \centering
  \small
  \caption{End-to-end latency breakdown of RhinoVLA on Huixi~R1.}
  \label{tab:e2e}
  \begin{tabular}{lcc}
    \toprule
    Stage & Latency (ms) & \% \\
    \midrule
    Vision Encoder (3 views)        & 24.31 & 28.4 \\
    VLM Backbone   & 20.78  & 24.3 \\
    Action Expert   & 36.71  & 42.9 \\
    Others   & 3.74  & 4.4 \\
    \midrule
    \textbf{Total}                           & 85.54  & 100.0 \\
    \textbf{Achieved closed-loop frequency}  & \multicolumn{2}{c}{\textbf{11.69 Hz}} \\
    \bottomrule
  \end{tabular}
\end{table}

\section{Conclusion}

We presented RhinoVLA, a deployment-oriented VLA model designed for real-time onboard robot control. By identifying VLM visual and context tokens as a major source of inference cost, RhinoVLA adopts a token-efficient Qwen3-VL backbone and an Action Expert to reduce VLM-side computation while preserving multimodal capability. To support cross-robot learning, RhinoVLA introduces a unified interface with view registry, a 72D physical state-action slot space, and robot-instance LoRA, enabling shared policy learning with embodiment-specific residual adaptation. With hardware-aware compilation, mixed-precision execution, and parallel visual encoding on Huixi R1, RhinoVLA achieves task performance comparable to $\pi_{0.5}$ and reaches 11.69 Hz end-to-end inference. 

\section{Future work}

RhinoVLA currently demonstrates an initial robot--model--chip workflow on Huixi R1, where the R1 platform is used for data collection and onboard deployment. In future open-source releases, we plan to extend this workflow by adding training support on R1, enabling the full data-collection, training, and deployment loop to run on the same edge SoC. This will make it possible to connect RhinoVLA with reinforcement learning pipelines and support real-robot online policy improvement directly on the deployed platform.

We will continue to follow more efficient and capable VLM backbones and update the visual-language module of RhinoVLA accordingly, such as future Qwen-series models, to further improve VLA performance while maintaining deployment efficiency.



\appendix

\newpage

\section{Additional VLA introduction}
\subsection{VLM Backbones in Recent VLA Models}

\begin{table}[ht]
\centering
\scriptsize
\begin{tabular}{p{0.18\linewidth} p{0.22\linewidth} p{0.50\linewidth}}
\hline
\textbf{VLM family} & \textbf{Representative use} & \textbf{Backbone characteristics} \\
\hline
PaLI-X / PaLM-E family~\cite{chen2023pali,driess2023palm} &
RT-2 &
Large web-trained VLMs; RT-2 casts robot actions into text-token form so robot
data can be co-trained with vision-language data. \\
Prismatic VLM: DINOv2 + SigLIP + Llama 2 &
OpenVLA &
Open 7B stack that fuses pretrained DINOv2 and SigLIP visual features with a
Llama 2 language model. \\
PaliGemma &
$\pi_{0}$, $\pi_{0.5}$ &
Compact open VLM based on SigLIP and Gemma; used as the visual-language prefix
before the $\pi$ action expert. \\
Eagle-2~\cite{li2025eagle} &
GR00T N1 &
VLM module paired with a diffusion-transformer action module in a dual-system
humanoid policy. \\
Qwen2-VL &
DexVLA &
Native dynamic resolution and M-RoPE for image, video, and text positions. \\
Qwen2.5-VL &
LingBot-VLA &
Native dynamic-resolution ViT, window attention, stronger grounding, structured
output, and absolute time encoding for video. \\
Qwen3-VL &
VLA Foundry; InternVLA-A1~\cite{cai2026internvla} &
Interleaved-MRoPE, DeepStack multi-level visual features, long interleaved
multimodal context, and text-timestamp alignment. \\
\hline
\end{tabular}
\caption{Representative VLM backbones used by, or relevant to, recent VLA
models. The table summarizes backbone characteristics only.}
\label{tab:vla_vlm_backbone_summary}
\end{table}

As summarized in Table~\ref{tab:vla_vlm_backbone_summary}, recent VLA systems increasingly build upon powerful pretrained VLM backbones rather than training perception modules from scratch. Early approaches such as RT-2 leverage large web-scale VLMs to transfer vision-language knowledge into robotic control, while OpenVLA adopts an open-source stack combining DINOv2, SigLIP, and Llama 2. More recent VLAs predominantly rely on the PaliGemma and Qwen-VL families. PaliGemma is used in $\pi_{0}$ and $\pi_{0.5}$ due to its compact design and strong vision-language capabilities, whereas the Qwen-VL series introduces dynamic-resolution processing, improved grounding, long-context multimodal reasoning, and structured output generation. These advances make modern VLM backbones increasingly suitable for embodied decision making, providing stronger perception, language understanding, and multimodal reasoning capabilities for downstream action generation.

\subsection{Rhino VLA Comparison with $\pi_{0.5}$}

\begin{table}[ht]
\centering
\scriptsize
\begin{tabular}{p{0.24\linewidth} p{0.31\linewidth} p{0.31\linewidth}}
\hline
\textbf{Item} & \textbf{RhinoVLA} & \textbf{$\pi_{0.5}$ reference} \\
\hline
\multicolumn{3}{l}{\textit{VLM side}} \\
VLM backbone &
Qwen3-VL-2B &
PaliGemma \\
VLM inference-path parameters &
2.13B &
2.92B \\
VLM text transformer &
28 layers, hidden size 2048, MLP hidden size 6144 &
18 layers, hidden size 2048, MLP hidden size 16384 \\
VLM text attention &
16 attention heads / 8 KV heads, head dimension 128 &
8 attention heads / 1 KV head, head dimension 256 \\
VLM vision transformer &
24 layers, hidden size 1024, MLP hidden size 4096, 16 heads &
27 layers, hidden size 1152, MLP hidden size 4304, 16 heads \\
Visual-token interface in the measured setting &
Images are resized to $256\times256$; each image gives a $16\times16$ raw
visual grid and 64 merged image tokens after spatial merging &
Fixed 256 image tokens per $224\times224$ image in the PaliGemma path \\
\hline
\multicolumn{3}{l}{\textit{Action Expert side}} \\
Action-side inference-path parameters &
0.40B &
0.43B \\
Action Expert depth / width &
18 layers / 1024 hidden size &
18 layers / 1024 hidden size \\
Action Expert MLP hidden size &
3072 &
4096 \\
Action Expert attention heads / KV heads &
16 / 8 &
8 / 1 \\
Action Expert GQA grouping &
2 query heads share one KV head &
8 query heads share one KV head \\
Action Expert head dimension &
128 &
256 \\
Action Expert attention computation &
Suffix tokens attend to Qwen3-VL prefix K/V and suffix K/V with Q/K
normalization and scaled dot-product attention &
Suffix tokens attend to the PaliGemma prefix context and suffix K/V in the
original $\pi$ action expert \\
Action-side tokens &
1 state token + 30 noisy action tokens &
30 noisy action tokens in our same-condition benchmark; released
$\pi_{0.5}$ uses $H=50$ \\
Visual-language context used by Action Expert &
Last 18 layers of Qwen3-VL KV cache &
PaliGemma prefix context used by the $\pi$ action expert \\
Action Expert block design &
Qwen-compatible attention/cache handling with AdaRMS conditioning and
Qwen-style MLP blocks &
Gemma expert blocks used by the original $\pi_{0.5}$ action model \\
Action interface &
Masked 72D physical state--action contract &
Reference fixed-dimensional action interface \\
\hline
\end{tabular}
\caption{Detailed architectural comparison between RhinoVLA and the
PaliGemma-based $\pi_{0.5}$ VLA. The comparison separates the VLM backbone from
the action side and reports the attention layout, GQA setting, hidden width,
and action-token interface used by each model. Parameter counts follow the
inference path: tied language embeddings are counted in the VLM backbone, while
unused output heads are excluded from the action-side count.}
\label{tab:rhino_vla_pi05_arch_detail}
\end{table}

As shown in Table~\ref{tab:rhino_vla_pi05_arch_detail}, Rhino VLA keeps the
Action Expert at a scale comparable to the $\pi_{0.5}$ reference while using a
more compact Qwen3-VL backbone. The action side also follows a
Qwen-compatible attention and cache layout, and extends the action interface to
the masked 72D physical slot space used for cross-robot training.
\section{Additional Details of the Unified 72D Action Interface}
\label{app:slot_mapping_details}

\subsection{Semantic Mapping Rule}
The unified 72D slot space is mapped semantically rather than positionally.
A raw joint or command is written to a slot only when its physical role matches
the slot definition. If a robot lacks a canonical degree of freedom, the
corresponding slot remains masked, and later joints are not shifted into the
empty position. This rule prevents an absent joint from changing the physical
meaning of subsequent action dimensions.

For example, a single-arm robot writes its arm joints to D0--D6 when their
roles match the canonical arm chain. A bimanual robot writes the left or first
arm to D0--D6 and the right or second arm to D7--D13. If a robot exposes fewer
than seven active arm joints, the missing canonical slots are masked rather
than filled by unrelated joints.

\subsection{State and Action Conversion}
State and action values are converted into the slot-defined physical units
whenever calibration and metadata are available. Joint angles are represented
in radians. Parallel grippers are represented as a closed ratio in $[0,1]$,
where 0 denotes fully open and 1 denotes fully closed. Mobile-base commands use
metric velocity units, including linear velocities in m/s and yaw rate in
rad/s.

When a dataset provides normalized, delta, or controller-specific commands, the
converter first checks whether the corresponding physical unit, joint limit,
and control convention are available. If the value can be reliably converted
into the slot-defined representation, it is written to the corresponding slot.
Otherwise, the value is kept outside the supervised 72D action target and the
slot is masked during training.

\subsection{Dexterous-Hand Slot Allocation}
Dexterous hands use a 16D active-control superset per hand. The first hand
uses H0--H15, corresponding to D16--D31, and the second hand uses H0--H15,
corresponding to D32--D47. The allocation follows a 4-3-3-3-3 structure from
thumb to little finger, as shown in
Table~\ref{tab:rhynix_vla_hand_slots_appendix}.

\begin{table}[t]
\centering
\scriptsize
\begin{tabular}{p{0.14\linewidth} p{0.19\linewidth} p{0.57\linewidth}}
\hline
\textbf{Per-hand slots} & \textbf{Finger} & \textbf{Meaning} \\
\hline
H0 & Thumb & Abduction/adduction \\
H1 & Thumb & Opposition rotation \\
H2 & Thumb & Metacarpophalangeal flexion/extension \\
H3 & Thumb & Interphalangeal flexion/extension \\
H4, H7, H10, H13 & Index, middle, ring, little & Abduction/adduction \\
H5, H8, H11, H14 & Index, middle, ring, little & Metacarpophalangeal flexion/extension \\
H6, H9, H12, H15 & Index, middle, ring, little & Proximal interphalangeal flexion/extension \\
\hline
\end{tabular}
\caption{Canonical per-hand allocation inside the unified 72D slot space. For
the first hand, H0--H15 correspond to D16--D31; for the second hand, they
correspond to D32--D47. Missing fingers or inactive DoF are masked rather than
reassigned to another finger.}
\label{tab:rhynix_vla_hand_slots_appendix}
\end{table}

This hand schema does not assume that every dexterous hand has 16 independent
motors. Only active, motor-controlled joints are supervised as action
dimensions. Passive joints, mechanically coupled follower joints, and
kinematically derived joints are not treated as separate action targets.
Lower-DoF hands map verified actuators to the closest canonical finger slots
and mask the remaining dimensions.

\subsection{Converter Constraints for Hand Commands}
A converter may write a source hand command into the 72D action target only
when the hardware or SDK documentation establishes the active-joint order,
unit, and joint limit. If a dataset provides normalized hand commands, the
converter maps them to nominal radians using hand-specific joint limits when
such limits are available. If the normalization convention or physical limits
are unknown, the corresponding values are not supervised as 72D action slots.

This conservative rule avoids assigning physically ambiguous hand commands to
fixed semantic slots. It also prevents passive or coupled hand motions from
being incorrectly interpreted as independently controllable action dimensions.

\subsection{Reserved Slots}
Slots D61--D71 are reserved auxiliary slots. They are inactive in the current
training schema and do not carry fixed semantics. These slots can only be
activated in future versions after their physical meanings, units, and masking
rules are explicitly defined. Until then, they remain masked in both model
inputs and training targets.

\section{Evaluation Details on Real Robots}
\label{sec:Evaluation}
Pick and place. During evaluation, objects are initialized at different tabletop positions within the task workspace. A rollout is considered successful only when the robot grasps the specified target object and places it into the specified target location. Grasping an incorrect object, placing the object into the wrong cell or region, dropping the object, or exceeding the rollout budget is counted as a failure.

Towel folding. During evaluation, the towel is initialized at different tabletop positions within the task workspace. A rollout is considered successful only when the robot lifts the towel with both arms and folds it in half. Failure cases include failing to grasp or lift the towel, dropping it during execution, producing an incomplete or incorrect fold, or exceeding the rollout budget.

\section{Author List}
The authors are listed in alphabetical order by first name: 

Chen Zhang, Chenyang Zhou, Guanghui He, Guanglei Ding, Haibin Gao, Jiajia Chen, Jianyong Zhang, Lianyi Yu, Ningyi Xu, Ping Xu, Qingchen Li, Quansheng Li, Yijia Zhang, Yingjun Hu, and Yuxi Liu

Chaochao Zhang, Chaoyi Li, Kai Liu, Kaikun Yang, and Zhichao Zhang
\section{Acknowledgments}
Kai Liu, Zhichao Zhang are from 21C LAB, Contemporary Amperex Technology Co., Limited, Ningde, China

Chaochao Zhang, Chaoyi Li are from Industrial Engineering Department, Contemporary Amperex Technology Co., Limited, Ningde, China

Kaikun Yang is from Department of After-market Business, Contemporary Amperex Technology Co., Limited, Ningde, China

\bibliographystyle{plain}
\bibliography{refs}

@article{brohan2022rt1,
  title        = {RT-1: Robotics Transformer for Real-World Control at Scale},
  author       = {Brohan, Anthony and Brown, Noah and Carbajal, Justice and Chebotar, Yevgen and others},
  journal      = {arXiv preprint arXiv:2212.06817},
  year         = {2022},
  url          = {https://arxiv.org/abs/2212.06817}
}

@inproceedings{zitkovich2023rt2,
  title        = {RT-2: Vision-Language-Action Models Transfer Web Knowledge to Robotic Control},
  author       = {Zitkovich, Brianna and Yu, Tianhe and Xu, Sichun and Xu, Peng and others},
  booktitle    = {Proceedings of The 7th Conference on Robot Learning},
  series       = {Proceedings of Machine Learning Research},
  volume       = {229},
  pages        = {2165--2183},
  year         = {2023},
  publisher    = {PMLR},
  url          = {https://proceedings.mlr.press/v229/zitkovich23a.html}
}

@inproceedings{kim2025openvla,
  title        = {OpenVLA: An Open-Source Vision-Language-Action Model},
  author       = {Kim, Moo Jin and Pertsch, Karl and Karamcheti, Siddharth and others},
  booktitle    = {Proceedings of The 8th Conference on Robot Learning},
  series       = {Proceedings of Machine Learning Research},
  year         = {2025},
  url          = {https://proceedings.mlr.press/v270/kim25c.html}
}

@article{black2024pi0,
  title        = {$\pi_0$: A Vision-Language-Action Flow Model for General Robot Control},
  author       = {Black, Kevin and Brown, Noah and Driess, Danny and Esmail, Adnan and others},
  journal      = {arXiv preprint arXiv:2410.24164},
  year         = {2024},
  url          = {https://arxiv.org/abs/2410.24164}
}

@article{bjorck2025grootn1,
  title        = {GR00T N1: An Open Foundation Model for Generalist Humanoid Robots},
  author       = {Bjorck, Johan and Castaneda, Fernando and Cherniadev, Nikita and Da, Xingye and others},
  journal      = {arXiv preprint arXiv:2503.14734},
  year         = {2025},
  url          = {https://arxiv.org/abs/2503.14734}
}

@misc{pi05technicalreport,
  title        = {$\pi_{0.5}$: A Vision-Language-Action Model with Open-World Generalization},
  author       = {{Physical Intelligence}},
  year         = {2025},
  howpublished = {Technical report},
  url          = {https://www.pi.website/download/pi05.pdf}
}

@misc{nvidia2022jetsonorin,
  title        = {NVIDIA Jetson AGX Orin Series Technical Brief},
  author       = {{NVIDIA}},
  year         = {2022},
  howpublished = {Technical brief},
  url          = {https://www.nvidia.com/content/dam/en-zz/Solutions/gtcf21/jetson-orin/nvidia-jetson-agx-orin-technical-brief.pdf}
}

@inproceedings{radford2021clip,
  title        = {Learning Transferable Visual Models From Natural Language Supervision},
  author       = {Radford, Alec and Kim, Jong Wook and Hallacy, Chris and Ramesh, Aditya and others},
  booktitle    = {Proceedings of the 38th International Conference on Machine Learning},
  series       = {Proceedings of Machine Learning Research},
  volume       = {139},
  pages        = {8748--8763},
  year         = {2021},
  publisher    = {PMLR},
  url          = {https://proceedings.mlr.press/v139/radford21a.html}
}

@article{liu2023llava,
  title        = {Visual Instruction Tuning},
  author       = {Liu, Haotian and Li, Chunyuan and Wu, Qingyang and Lee, Yong Jae},
  journal      = {arXiv preprint arXiv:2304.08485},
  year         = {2023},
  url          = {https://arxiv.org/abs/2304.08485}
}

@article{beyer2024paligemma,
  title        = {PaliGemma: A Versatile 3B VLM for Transfer},
  author       = {Beyer, Lucas and Steiner, Andreas and Pinto, Andre Susano and Kolesnikov, Alexander and others},
  journal      = {arXiv preprint arXiv:2407.07726},
  year         = {2024},
  url          = {https://arxiv.org/abs/2407.07726}
}

@article{wang2024qwen2vl,
  title        = {Qwen2-VL: Enhancing Vision-Language Model's Perception of the World at Any Resolution},
  author       = {Wang, Peng and Bai, Shuai and Tan, Sinan and others},
  journal      = {arXiv preprint arXiv:2409.12191},
  year         = {2024},
  url          = {https://arxiv.org/abs/2409.12191}
}

@article{qwen2025qwen25vl,
  title        = {Qwen2.5-VL Technical Report},
  author       = {{Qwen Team}},
  journal      = {arXiv preprint arXiv:2502.13923},
  year         = {2025},
  url          = {https://arxiv.org/abs/2502.13923}
}

@article{bai2025qwen3vl,
  title        = {Qwen3-VL Technical Report},
  author       = {Bai, Shuai and Cai, Yuxuan and others},
  journal      = {arXiv preprint arXiv:2511.21631},
  year         = {2025},
  url          = {https://arxiv.org/abs/2511.21631}
}

@article{openx2023rtx,
  title        = {Open X-Embodiment: Robotic Learning Datasets and RT-X Models},
  author       = {O'Neill, Abby and Rehman, Abdul and Gupta, Abhishek and Maddukuri, Abhiram and others},
  journal      = {arXiv preprint arXiv:2310.08864},
  year         = {2023},
  url          = {https://arxiv.org/abs/2310.08864}
}

@article{liu2024rdt,
  title        = {RDT-1B: A Diffusion Foundation Model for Bimanual Manipulation},
  author       = {Liu, Songming and Wu, Lingxuan and Li, Bangguo and Tan, Hengkai and others},
  journal      = {arXiv preprint arXiv:2410.07864},
  year         = {2024},
  url          = {https://arxiv.org/abs/2410.07864}
}

@inproceedings{hu2022lora,
  title        = {LoRA: Low-Rank Adaptation of Large Language Models},
  author       = {Hu, Edward J. and Shen, Yelong and Wallis, Phillip and Allen-Zhu, Zeyuan and others},
  booktitle    = {International Conference on Learning Representations},
  year         = {2022},
  url          = {https://openreview.net/forum?id=nZeVKeeFYf9}
}

@article{luo2024hilserl,
  title        = {Precise and Dexterous Robotic Manipulation via Human-in-the-Loop Reinforcement Learning},
  author       = {Luo, Jianlan and others},
  journal      = {arXiv preprint arXiv:2410.21845},
  year         = {2024},
  url          = {https://arxiv.org/abs/2410.21845}
}

@inproceedings{jang2022bcz,
  title        = {BC-Z: Zero-Shot Task Generalization with Robotic Imitation Learning},
  author       = {Jang, Eric and Irpan, Alex and Khansari, Mohi and Kappler, Daniel and others},
  booktitle    = {Proceedings of the 5th Conference on Robot Learning},
  series       = {Proceedings of Machine Learning Research},
  volume       = {164},
  pages        = {991--1002},
  year         = {2022},
  publisher    = {PMLR},
  url          = {https://proceedings.mlr.press/v164/jang22a.html}
}

@inproceedings{peebles2023scalable,
  title={Scalable diffusion models with transformers},
  author={Peebles, William and Xie, Saining},
  booktitle={Proceedings of the IEEE/CVF international conference on computer vision},
  pages={4195--4205},
  year={2023}
}

@article{oquab2023dinov2,
  title={Dinov2: Learning robust visual features without supervision},
  author={Oquab, Maxime and Darcet, Timoth{\'e}e and Moutakanni, Th{\'e}o and Vo, Huy and Szafraniec, Marc and Khalidov, Vasil and Fernandez, Pierre and Haziza, Daniel and Massa, Francisco and El-Nouby, Alaaeldin and others},
  journal={arXiv preprint arXiv:2304.07193},
  year={2023}
}

@inproceedings{zhai2023sigmoid,
  title={Sigmoid loss for language image pre-training},
  author={Zhai, Xiaohua and Mustafa, Basil and Kolesnikov, Alexander and Beyer, Lucas},
  booktitle={Proceedings of the IEEE/CVF international conference on computer vision},
  pages={11975--11986},
  year={2023}
}

@article{touvron2023llama,
  title={Llama 2: Open foundation and fine-tuned chat models},
  author={Touvron, Hugo and Martin, Louis and Stone, Kevin and Albert, Peter and Almahairi, Amjad and Babaei, Yasmine and Bashlykov, Nikolay and Batra, Soumya and Bhargava, Prajjwal and Bhosale, Shruti and others},
  journal={arXiv preprint arXiv:2307.09288},
  year={2023}
}

@article{wen2025dexvla,
  title={Dexvla: Vision-language model with plug-in diffusion expert for general robot control},
  author={Wen, Junjie and Zhu, Yichen and Li, Jinming and Tang, Zhibin and Shen, Chaomin and Feng, Feifei},
  journal={arXiv preprint arXiv:2502.05855},
  year={2025}
}

@article{wu2026pragmatic,
  title={A Pragmatic VLA Foundation Model},
  author={Wu, Wei and Lu, Fan and Wang, Yunnan and Yang, Shuai and Liu, Shi and Wang, Fangjing and Zhu, Qian and Sun, He and Wang, Yong and Ma, Shuailei and others},
  journal={arXiv preprint arXiv:2601.18692},
  year={2026}
}

@misc{nvidia_geforce_rtx4090_specs,
  title        = {{GeForce RTX 4090}: Graphics Cards for Gaming},
  author       = {{NVIDIA}},
  year         = {2022},
  howpublished = {Official product specification page},
  url          = {https://www.nvidia.com/en-sg/geforce/graphics-cards/40-series/rtx-4090/},
  note         = {Accessed: 2026-06-02}
}

@misc{nvidia_geforce_rtx5090_specs,
  title        = {{GeForce RTX 5090}: Graphics Cards for Gamers and Creators},
  author       = {{NVIDIA}},
  year         = {2025},
  howpublished = {Official product specification page},
  url          = {https://www.nvidia.com/en-sg/geforce/graphics-cards/50-series/rtx-5090/},
  note         = {Accessed: 2026-06-02}
}

@misc{nvidia2025jetsonthor_datasheet,
  title        = {NVIDIA Jetson Thor Series Modules Data Sheet},
  author       = {{NVIDIA Corporation}},
  year         = {2025},
  howpublished = {Official Datasheet},
  url          = {https://developer.nvidia.com/downloads/assets/embedded/secure/jetson/thor/docs/jetson-thor-series-modules-datasheet_ds-11945-001.pdf},
  note         = {Accessed: 2026-06-02}
}

@article{bu2025agibotworld,
  title={AgiBot World Colosseo: A Large-scale Manipulation Platform for Scalable and Intelligent Embodied Systems},
  author={Bu, Qingwen and Cai, Jisong and Chen, Li and Cui, Xiuqi and Ding, Yan and Feng, Siyuan and Gao, Shenyuan and He, Xindong and Huang, Xu and Jiang, Shu and others},
  journal={arXiv preprint arXiv:2503.06669},
  year={2025}
}

@article{dao2022flashattention,
  title={Flashattention: Fast and memory-efficient exact attention with io-awareness},
  author={Dao, Tri and Fu, Dan and Ermon, Stefano and Rudra, Atri and R{\'e}, Christopher},
  journal={Advances in neural information processing systems},
  volume={35},
  pages={16344--16359},
  year={2022}
}

@article{zhang2019root,
  title={Root mean square layer normalization},
  author={Zhang, Biao and Sennrich, Rico},
  journal={Advances in neural information processing systems},
  volume={32},
  year={2019}
}

@article{chen2023pali,
  title={Pali-x: On scaling up a multilingual vision and language model},
  author={Chen, Xi and Djolonga, Josip and Padlewski, Piotr and Mustafa, Basil and Changpinyo, Soravit and Wu, Jialin and Ruiz, Carlos Riquelme and Goodman, Sebastian and Wang, Xiao and Tay, Yi and others},
  journal={arXiv preprint arXiv:2305.18565},
  year={2023}
}

@article{driess2023palm,
  title={Palm-e: An embodied multimodal language model},
  author={Driess, Danny and Xia, Fei and Sajjadi, Mehdi SM and Lynch, Corey and Chowdhery, Aakanksha and Ichter, Brian and Wahid, Ayzaan and Tompson, Jonathan and Vuong, Quan and Yu, Tianhe and others},
  journal={arXiv preprint arXiv:2303.03378},
  year={2023}
}

@article{li2025eagle,
  title={Eagle 2: Building post-training data strategies from scratch for frontier vision-language models},
  author={Li, Zhiqi and Chen, Guo and Liu, Shilong and Wang, Shihao and VS, Vibashan and Ji, Yishen and Lan, Shiyi and Zhang, Hao and Zhao, Yilin and Radhakrishnan, Subhashree and others},
  journal={arXiv preprint arXiv:2501.14818},
  year={2025}
}

@article{cai2026internvla,
  title={InternVLA-A1: Unifying Understanding, Generation and Action for Robotic Manipulation},
  author={Cai, Junhao and Cai, Zetao and Cao, Jiafei and Chen, Yilun and He, Zeyu and Jiang, Lei and Li, Hang and Li, Hengjie and Li, Yang and Liu, Yufei and others},
  journal={arXiv preprint arXiv:2601.02456},
  year={2026}
}

@article{chi2025diffusion,
  title        = {Diffusion Policy: Visuomotor Policy Learning via Action Diffusion},
  author       = {Chi, Cheng and Xu, Zhenjia and Feng, Siyuan and Cousineau, Eric and Du, Yilun and Burchfiel, Benjamin and Tedrake, Russ and Song, Shuran},
  journal      = {The International Journal of Robotics Research},
  volume       = {44},
  number       = {10--11},
  year         = {2025},
  doi          = {10.1177/02783649241273668},
  url          = {https://doi.org/10.1177/02783649241273668}
}

@inproceedings{octo2024,
  title        = {Octo: An Open-Source Generalist Robot Policy},
  author       = {{Octo Model Team} and Ghosh, Dibya and Walke, Homer and Pertsch, Karl and Black, Kevin and Mees, Oier and Dasari, Sudeep and Hejna, Joey and Kreiman, Tobias and Xu, Charles and Luo, Jianlan and Tan, You Liang and Chen, Lawrence Yunliang and Sanketi, Pannag and Vuong, Quan and Xiao, Ted and Sadigh, Dorsa and Finn, Chelsea and Levine, Sergey},
  booktitle    = {Robotics: Science and Systems},
  year         = {2024},
  url          = {https://arxiv.org/abs/2405.12213}
}

@inproceedings{reuss2024mdt,
  title        = {Multimodal Diffusion Transformer: Learning Versatile Behavior from Multimodal Goals},
  author       = {Reuss, Moritz and Ya{\u{g}}murlu, {\"O}mer Erdin{\c{c}} and Wenzel, Fabian and Lioutikov, Rudolf},
  booktitle    = {Robotics: Science and Systems},
  year         = {2024},
  url          = {https://arxiv.org/abs/2407.05996}
}

@inproceedings{zheng2025tracevla,
  title        = {TraceVLA: Visual Trace Prompting Enhances Spatial-Temporal Awareness for Generalist Robotic Policies},
  author       = {Zheng, Ruijie and Liang, Yongyuan and Huang, Shuaiyi and Gao, Jianfeng and Daum{\'e} III, Hal and Kolobov, Andrey and Huang, Furong and Yang, Jianwei},
  booktitle    = {International Conference on Learning Representations},
  year         = {2025},
  url          = {https://openreview.net/forum?id=b1CVu9l5GO}
}

@inproceedings{qu2025spatialvla,
  title        = {SpatialVLA: Exploring Spatial Representations for Visual-Language-Action Model},
  author       = {Qu, Delin and Song, Haoming and Chen, Qizhi and Yao, Yuanqi and Ye, Xinyi and Ding, Yan and Wang, Zhigang and Gu, Jiayuan and Zhao, Bin and Wang, Dong and Li, Xuelong},
  booktitle    = {Robotics: Science and Systems},
  year         = {2025},
  url          = {https://arxiv.org/abs/2501.15830}
}

@article{cen2025worldvla,
  title        = {WorldVLA: Towards Autoregressive Action World Model},
  author       = {Cen, Jun and Yu, Chaohui and Yuan, Hangjie and Jiang, Yuming and Huang, Siteng and Guo, Jiayan and Li, Xin and Song, Yibing and Luo, Hao and Wang, Fan and Zhao, Deli and Chen, Hao},
  journal      = {arXiv preprint arXiv:2506.21539},
  year         = {2025},
  url          = {https://arxiv.org/abs/2506.21539}
}

@inproceedings{zhao2025cotvla,
  title        = {CoT-VLA: Visual Chain-of-Thought Reasoning for Vision-Language-Action Models},
  author       = {Zhao, Qingqing and Lu, Yao and Kim, Moo Jin and Fu, Zipeng and Zhang, Zhuoyang and Wu, Yecheng and Li, Zhaoshuo and Ma, Qianli and Han, Song and Finn, Chelsea and Handa, Ankur and Liu, Ming-Yu and Xiang, Donglai and Wetzstein, Gordon and Lin, Tsung-Yi},
  booktitle    = {Proceedings of the IEEE/CVF Conference on Computer Vision and Pattern Recognition},
  year         = {2025},
  url          = {https://arxiv.org/abs/2503.22020}
}

@article{pertsch2025fast,
  title        = {FAST: Efficient Action Tokenization for Vision-Language-Action Models},
  author       = {Pertsch, Karl and Stachowicz, Kyle and Ichter, Brian and Driess, Danny and Nair, Suraj and Vuong, Quan and Mees, Oier and Finn, Chelsea and Levine, Sergey},
  journal      = {arXiv preprint arXiv:2501.09747},
  year         = {2025},
  url          = {https://arxiv.org/abs/2501.09747}
}

@article{hung2025nora,
  title        = {NORA: A Small Open-Sourced Generalist Vision Language Action Model for Embodied Tasks},
  author       = {Hung, Chia-Yu and Sun, Qi and Hong, Pengfei and Zadeh, Amir and Li, Chuan and Tan, U-Xuan and Majumder, Navonil and Poria, Soujanya},
  journal      = {arXiv preprint arXiv:2504.19854},
  year         = {2025},
  url          = {https://arxiv.org/abs/2504.19854}
}

@article{shukor2025smolvla,
  title        = {SmolVLA: A Vision-Language-Action Model for Affordable and Efficient Robotics},
  author       = {Shukor, Mustafa and Aubakirova, Dana and Capuano, Francesco and Kooijmans, Pepijn and Palma, Steven and Zouitine, Adil and Aractingi, Michel and Pascal, Caroline and Russi, Martino and Marafioti, Andres and Alibert, Simon and Cord, Matthieu and Wolf, Thomas and Cadene, Remi},
  journal      = {arXiv preprint arXiv:2506.01844},
  year         = {2025},
  url          = {https://arxiv.org/abs/2506.01844}
}

@misc{agibotworld2026,
    title        = {AgiBot World 2026},
    author       = {AgiBot World Team},
    howpublished = {\url{https://huggingface.co/datasets/agibot-world/AgiBotWorld2026}},
    year         = {2026}
}

@misc{contributors2024agibotworldrepo,
  title={AgiBot World Colosseum},
  author={AgiBot World Colosseum contributors},
  howpublished={\url{https://github.com/OpenDriveLab/AgiBot-World}},
  year={2024}
}

\end{document}